\documentclass{article}

\usepackage[preprint]{neurips_2021}

\usepackage{hyperref}  
\hypersetup{
    colorlinks=true,
    linkcolor=blue,
    citecolor =blue,
    filecolor=magenta,      
    urlcolor=magenta,
}

\usepackage[utf8]{inputenc} % allow utf-8 input
\usepackage[T1]{fontenc}    % use 8-bit T1 fonts
\usepackage{hyperref}       % hyperlinks
\usepackage{url}            % simple URL typesetting
\usepackage{booktabs}       % professional-quality tables
\usepackage{amsfonts}       % blackboard math symbols
\usepackage{nicefrac}       % compact symbols for 1/2, etc.
\usepackage{microtype}      % microtypography
\usepackage{xcolor}         % colors

\title{Optimal Order Simple Regret for Gaussian Process Bandits}

\author{%
Sattar Vakili$^*$, Nacime Bouziani$^\dagger$, Sepehr Jalali$^*$, Alberto Bernacchia$^*$, Da-shan Shiu$^*$ \\
$^*$ MediaTek Research\\
\texttt{\{sattar.vakili, sepehr.jalali, alberto.bernacchia, ds.shiu\}@mtkresearch.com} \\
$^\dagger$Imperial College London\\
\texttt{n.bouziani18@imperial.ac.uk } \\
}

\usepackage{bm}
\usepackage{amssymb}
\usepackage{amsmath}

\def\tcb{\textcolor{blue}}

\def\nn{\nonumber}
\def\TP{\top}
\def\argmax{\text{argmax}}
\def\argmin{\text{argmin}}

\def\Rr{\mathbb{R}}
\def\Nn{\mathbb{N}}
\def\E{\mathbb{E}} %%%

\def\Hc{\mathcal{H}}

\def\Oc{\mathcal{O}}
\def\Xc{\mathcal{X}}
\def\Ac{\mathcal{A}}
\def\Sc{\mathcal{S}}
\def\Dc{\mathcal{D}}
\def\Ic{\mathcal{I}}
\def\Ec{\mathcal{E}}

\newtheorem{lemma}{Lemma}
\newtheorem{theorem}{Theorem}
\newtheorem{proposition}{Proposition}

\newtheorem{definition}{Definition}
\newtheorem{corollary}{Corollary}
\newtheorem{remark}{Remark}

\newtheorem{assumption}{Assumption}

\usepackage{float}
\usepackage{cleveref}
\crefname{section}{§}{§§}
\Crefname{section}{§}{§§}
\usepackage{makecell}
\usepackage{graphicx} % demo is just for the example
\usepackage{subfig}
\usepackage{algpseudocode}
\usepackage{algorithm}
\begin{document}

\maketitle

\begin{abstract}
  Consider the sequential optimization of a continuous, possibly non-convex, and expensive to evaluate objective function $f$. The problem can be cast as a Gaussian Process (GP) bandit where $f$ lives in a reproducing kernel Hilbert space (RKHS). The state of the art analysis of several learning algorithms shows a significant gap between the lower and upper bounds on the simple regret performance. When $N$ is the number of exploration trials and $\gamma_N$ is the maximal information gain, we prove an $\tilde{\Oc}(\sqrt{\gamma_N/N})$ bound on the simple regret performance of a pure exploration algorithm that is significantly tighter than the existing bounds. We show that this bound is order optimal up to logarithmic factors for the cases where a lower bound on regret is known. To establish these results, we prove novel and sharp confidence intervals for GP models applicable to RKHS elements which may be of broader interest. 
\end{abstract}

\section{Introduction}\label{sec:Intro}

Sequential optimization has evolved into one of the fastest developing areas of machine
learning~\citep{mazumdar2020langevinn}. We consider sequential optimization of an unknown objective function from noisy and expensive to evaluate zeroth-order\footnote{Zeroth-order feedback signifies observations from $f$ in contrast to first-order feedback which refers to observations from gradient of $f$ as e.g. in stochastic gradient descent~ \citep[see, e.g.,][]{Agrawal2011, VakiliISIT2019}.} observations.
That is a ubiquitous problem in academic research and industrial production. Examples of applications include exploration in reinforcement learning, recommendation systems, medical analysis tools and speech recognizers \citep[][]{Shahriari2016outofloop}.
% Another notable application is hyper-parameter tuning in machine learning models where commonly used methods such as random or grid search are prohibitively expensive~\citep[see e.g.][]{bergstra2011algorithms, mcgibbon2016osprey}.
%Sequential optimization methods on the other hand are shown to be able to efficiently find good configurations of hyper-parameters by adaptively exploring the hyper-parameter space based on the knowledge acquired through past observations~\cite{falkner2018bohb}.
A notable application in the field of machine learning is automatic hyper-parameter tuning. Prevalent methods such as grid search can be prohibitively expensive~\citep[][]{bergstra2011algorithms, mcgibbon2016osprey}. Sequential optimization methods, on the other hand, are shown to efficiently find good hyper-parameters by an adaptive exploration of the hyper-parameter space~\citep{falkner2018bohb}.

% One approach to hyper-parameter tuning is random or grid search~\citep{mcgibbon2016osprey}, which is expensive and biased by the subjective choice of the practitioner on which part of the space to explore. 
% Another approach is gradient-based, in which derivatives are chained through the entire training procedure (Maclaurin, Duvenaud, Adams). This can be used only when gradients are available and is still very expensive to run. 
% Instead, sequential optimization is gradient-free and is able to quickly find good configurations by using the knowledge acquired by past observations to actively choose the next ones. 

%One might try to discover a set of high performance hyper-parameter configuration by running a sequential experiment. %The performance of the model is often a non-convex function of hyper-parameters. A learning algorithm is allowed to sequentially test the performance of the model on a sequence of hyper-parameter configurations where the goal is to return the hyper-parameter configuration with the best performance~\citep{falkner2018bohb}. 

Our sequential optimization setting is as follows. Consider an objective function $f$ defined over a domain $\Xc \subset \Rr^d$, where $d\in\Nn$ is the dimension of the input. A learning algorithm is allowed to perform an adaptive exploration to sequentially observe the potentially corrupted values of the objective function  $\{f(x_n)+\epsilon_n\}_{n=1}^N$, where $\epsilon_n$ are random noises. At the end of $N$ exploration trials, the learning algorithm returns a candidate maximizer $\hat{x}_N^*\in \Xc$ of $f$. Let $x^* \in \text{argmax}_{x\in\Xc} f(x)$ be a true optimal solution. We may measure the performance of the learning algorithm in terms of \textit{simple regret}; that is, the difference between the performance under the true optimal, $f(x^*)$, and that under the learnt value, $f(\hat{x}_N^*)$.

Our formulation falls under the general framework of continuum armed bandits that signifies receiving feedback only for the selected observation point $x_n$ at each time $n$~\citep{agrawal1995continuum, kleinberg2004nearly, bubeck2011pure, bubeck2011x}. 
Bandit problems have been extensively studied under numerous settings and various performance measures including simple regret~\citep[see, e.g.,][]{bubeck2011pure, carpentier2015simple, deshmukh2018simple}, cumulative regret~\citep[see, e.g.,][]{Aeur2002UCB, slivkins2019introduction,zhao2019book}, and best arm identification~\citep[see, e.g.,][]{audibert2010best, grover2018best}. 
The choice of performance measure strongly depends on the application.
Simple regret is suitable for situations with a preliminary exploration phase (for instance hyper-parameter tuning) in which costs are not measured in terms of rewards but rather in terms of resources expended~\citep{bubeck2011pure}. 

Due to infinite cardinality of the domain, approaching $f(x^*)$ is feasible only when appropriate regularity assumptions on $f$ and noise are satisfied. Following a growing literature~\citep[][]{srinivas2010gaussian, Chowdhury2017bandit, Janz2020SlightImprov, vakili2020information}, we focus on a variation of the problem where $f$ is assumed to belong to a reproducing kernel Hilbert space (RKHS) that is a very general assumption. Almost all continuous functions can be approximated with the RKHS elements of practically relevant kernels such as Mat{\'e}rn family of kernels~\citep{srinivas2010gaussian}. We consider two classes of noise: sub-Gaussian and light-tailed. 
%We prove the results under both assumptions on noise. 

Our regularity assumption on $f$ allows us to utilize
Gaussian processes (GPs) which provide powerful Bayesian (surrogate) models for $f$~\citep{Rasmussen2006}. Sequential optimization based on GP models is often referred to as Bayesian optimization in the literature~\citep[][]{ Shahriari2016outofloop,Snoek2012practicalBO, frazier2018bayesian}. We build on prediction and uncertainty estimates provided by GP models to study an efficient adaptive exploration algorithm referred to as Maximum Variance Reduction (MVR). Under simple regret measure, MVR embodies the simple principle of exploring the points with the highest variance first. Intuitively, the variance in the GP model 
%(what is the variance in the GP model? the variance learned by a GP model?)
is considered as a measure of uncertainty about the unknown objective function and the exploration steps are designed to maximally reduce the uncertainty. At the end of exploration trials, MVR returns a candidate maximizer based on the prediction provided by the learnt GP model. With its simple structure, MVR is amenable to a tight analysis that significantly improves the best known bounds on simple regret.
To this end, we derive novel and sharp confidence intervals for GP models applicable to RKHS elements. 
In addition, we provide numerical experiments on the simple regret performance of MVR comparing it to GP-UCB~\citep{srinivas2010gaussian, Chowdhury2017bandit}, GP-PI~\citep{hoffman2011portfolio} and GP-EI~\citep{hoffman2011portfolio}.

%(Consider moving this paragraph up, to the third paragraph of the introduction, if you can eliminate the "based on GP models" in this paragraph.) 
%The framework of sequential optimization based on GP models is often referred to as Bayesian optimization in the literature~(see e.g. \cite{Shahriari2016outofloop}).
%In addition to the hyper parameter tuning example given above, Bayesian optimization finds a verity of applications in e.g. best action exploration in reinforcement learning, recommendation systems, medical analysis tools, speech recognizers, and industrial production (see e.g.~\cite{Shahriari2016outofloop} and references therein). 

\subsection{Main Results}
Our main contributions are as follows. 

%\begin{itemize}
    %\item 
    % We first derive novel confidence intervals for GP models applicable to RKHS elements (Theorems~\ref{The:ConIneqSubG} and~\ref{The:ConIneqLT}). As part of our analysis, we prove an expression for the posterior variance of a GP model in terms of the maximum prediction error for an RKHS element from noise-free observations and the effect of noise (Proposition~\ref{Prop:var}) which draws new connections between GP regression and kernel ridge regression~\cite{Kanagawa2018}. These results may be of independent interest.  
    
We first derive novel confidence intervals for GP models applicable to RKHS elements (Theorems~\ref{The:ConIneqSubG} and~\ref{The:ConIneqLT}). As part of our analysis, we formulate the posterior variance of a GP model as the sum of two terms: the maximum prediction error from noise-free observations, and the effect of noise (Proposition~\ref{Prop:var}). This interpretation elicits new connections between GP regression and kernel ridge regression~\citep{Kanagawa2018}. These results are of interest on their own. 
    
%\item 
We then build on the confidence intervals for GP models to provide a tight analysis of the simple regret of the MVR algorithm (Theorem~\ref{The:regSubG}). 
%We build on the confidence intervals proven for the GP models, to provide a tight analysis of the simple regret for the MVR algorithm (Theorems~\ref{The:regSubG} and~\ref{The:regLT}).
In particular, we prove a high probability  $\tilde{\Oc}(\sqrt{\frac{\gamma_N}{N}})$\footnote{The notations $\Oc$ and $\tilde{\Oc}$ are used to denote the mathematical order and the mathematical order up to logarithmic factors, respectively.} simple regret, where $\gamma_N$ is the maximal information gain (see~\cref{Sec:InfoGain}). In comparison to the existing $\tilde{\Oc}(\frac{\gamma_N}{\sqrt{N}})$ bounds on simple regret~\citep[see, e.g.,][]{srinivas2010gaussian, Chowdhury2017bandit, Scarlett2017Lower}, we show an $\Oc(\sqrt{\gamma_N})$ improvement. 
It is noteworthy that our bound guarantees convergence to the optimum value of $f$, while previous $\tilde{\Oc}(\frac{\gamma_N}{\sqrt{N}})$ bounds do not, since although $\gamma_N$ grows sublinearly with $N$, it can grow faster than $\sqrt{N}$.
    
    %\item 
We then specialize our results for the particular cases of practically relevant Mat{\'e}rn and Squared Exponential (SE) kernels. We show that our regret bounds match the lower bounds and close the gap reported in~\cite{Scarlett2017Lower, cai2020lower}, who showed that an average simple regret of $\epsilon$ requires $N= \Omega\left(\frac{1}{\epsilon^2}(\log(\frac{1}{\epsilon}))^{\frac{d}{2}}\right)$ exploration trials in the case of SE kernel. For the Mat{\'e}rn-$\nu$ kernel (where $\nu$ is the smoothness parameter, see~\cref{Sec:GPs}) they gave the analogous bound of $N=\Omega\left((\frac{1}{\epsilon})^{2+\frac{d}{\nu}}\right)$. They also reported a significant gap between these lower bounds and the upper bounds achieved by GP-UCB algorithm. In Corollary~\ref{cor1}, we show that our analysis of MVR closes this gap in the performance and establishes upper bounds matching the lower bounds up to logarithmic factors.

%\item 
In contrast to the existing results which mainly focus on Gaussian and sub-Gaussian distributions for noise, we extend our analysis to the more general class of light-tailed distributions, thus broadening the applicability of the results. This extension increases both the confidence interval width and the simple regret by only a multiplicative logarithmic factor. These results apply to e.g. the privacy preserving setting where often a light-tailed noise is employed~\citep{basu2019differential,ren2020LDP,zheng2020LDP}.
    
%\item 
% We perform several experiments on synthetic and benchmark test functions comparing the performance of MVR in terms of simple regret with other Bayesian optimization algorithms such as GP-UCB, GP-PI and GP-EI. Our experiments show a better performance for MVR in terms of simple regret.
%\end{itemize}

\subsection{Literature Review}\label{sec:LitRef}

The celebrated work of Srinivas \emph{et al.} \cite{srinivas2010gaussian} pioneered the analysis of GP bandits by proving an $\tilde{\Oc}(\gamma_N\sqrt{N})$ upper bound on the cumulative regret of GP-UCB, an optimistic optimization algorithm which sequentially selects $x_n$ that maximize an upper confidence bound score over the search space. That implies an $\tilde{\Oc}(\frac{\gamma_N}{\sqrt{N}})$ simple regret~\citep{Scarlett2017Lower}. Their analysis relied on deriving confidence intervals for GP models applicable to RKHS elements. They also considered a fully Bayesian setting where $f$ is assumed to be a sample from a GP and noise is assumed to be Gaussian.  \cite{Chowdhury2017bandit} built on feature space representation of GP models and self-normalized martingale inequalities, first developed in~\cite{Abbasi2011} for linear bandits, to improve the confidence intervals of~\cite{srinivas2010gaussian} by a multiplicative $\log(N)$ factor. That led to an improvement in the regret bounds by the same multiplicative $\log(N)$ factor. A discussion on the comparison between these results and the confidence intervals derived in this paper is provided in~\cref{comp:conf}. A technical comparison with some recent advances in regret bounds requires introducing new notations and is deferred to~Appendix~\tcb{A}.
%In addition,~\cite{Chowdhury2017bandit} extended the analysis of GP-UCB to GP-TS, a Bayesian
%optimization algorithm based on Thompson Sampling which sequentially draws $x_n$ from the posterior distribution of $x^*$, and showed the same order of regret as GP-UCB for GP-TS. 

% \cite{Scarlett2017Lower} derived lower bounds on cumulative and simple regret for the GP Bandit problems. In particular, they showed that for the SE kernel, an average simple regret of $\epsilon$ requires $N= \Omega\left(\frac{1}{\epsilon^2}(\log(\frac{1}{\epsilon}))^{\frac{d}{2}}\right)$. For the Mat{\'e}rn-$\nu$ kernel where $\nu$ is the smoothness parameter (see~\cref{Sec:GPs}) they gave the analogous bound of $N=\Omega\left((\frac{1}{\epsilon})^{2+\frac{d}{\nu}}\right)$. While they reported 

The performance of Bayesian optimization algorithms has been extensively studied under numerous settings including contextual information
~\citep{Krause11Contexual}, high dimensional spaces~\citep{Josip13HighD, Mutny2018SGPTS}, safety constraints~\citep{berkenkamp2016bayesiansafe, sui2018stagewisesafe}, parallelization~\citep{kandasamy2018parallelised}, meta-learning~\citep{wang2018meta}, multi-fidelity evaluations~\citep{kandasamy2019multifidelity}, ordinal models~\citep{picheny2019ordinal}, corruption tolerance~\citep{bogunovic2020corruption, cai2020lower}, and neural tangent kernels~\citep{zhou2020neuralUCB, zhang2020neuralTS}.
\cite{Javidi} introduced an adaptive discretization of the search space improving the computational complexity of a GP-UCB based algorithm. Sparse approximation of GP posteriors are shown to preserve the regret orders while improving the computational complexity of Bayesian optimization algorithms~\citep{Mutny2018SGPTS, Calandriello2019Adaptive, Vakili2020Scalable}. Under the RKHS setting with noisy observations, GP-TS~\citep{Chowdhury2017bandit} and GP-EI~\citep{nguyen2017regretGPEI, wang2014theoreticalGPEI} are also shown to achieve the same regret guarantees as GP-UCB (up to logarithmic factors). All these works report $\tilde{\Oc}(\frac{\gamma_N}{\sqrt{N}})$ regret bounds. 

The regret bounds are also reported under other often simpler settings such as noise-free observations~\citep[][$\epsilon_n=0, \forall n$]{Bull2011EI, Vakili20NoiseFree} or a Bayesian regret that is averaged over a known prior on $f$~\citep{kandasamy2018parallelised, wang2018regretGPPI, wang2017, scarlett2018tight, shekhar2021significance, grunewalder2010regret, Freitas2012,  Kawaguchi2015}, rather than for a fixed and unknown $f$ as in our setting.

Other lines of work on continuum armed bandits exist relying on other regularity assumptions such as Lipschitz continuity~\citep{kleinberg2004nearly, bubeck2011x, carpentier2015simple, kleinberg2008multi}, convexity~\citep{Agrawal2011} and unimodality~\citep{combes2020unimodal}, to name a few. A notable example is~\cite{bubeck2011x} who showed that hierarchical algorithms based on tree search yield $\Oc(N^{\frac{d+1}{d+2}})$ cumulative regret. We do not compare with these results due to the inherent difference in the regularity assumptions. 

% Bandit problems have been extensively studied under other settings with different regularity assumptions than ours and various performance measures including cumulative regret~(see e.g~\cite{Aeur2002UCB, slivkins2019introduction,zhao2019book}), simple regret (see e.g.~\cite{bubeck2011pure, carpentier2015simple, deshmukh2018simple}), and best arm identification (e.g. see~\cite{audibert2010best, grover2018best}). 

%The rest of the paper is organized as follows. The problem formulation, regularity assumptions, the preliminaries on RKHS, GP models and information gain are presented in~\cref{sec:PF}. The novel confidence intervals for GP models are provided in~\cref{sec:ConfidenceIntervals}. MVR algorithm and its analysis are given in~\cref{Sec:SimpleR}. The experiments are presented in~\cref{Sec:Exp}. We conclude with a discussion in~\cref{Sec:Disc}.

\subsection{Organization}

In~\cref{sec:PF}, the problem formulation, the regularity assumptions, and the preliminaries on RKHS and GP models are presented. The novel confidence intervals for GP models are proven in~\cref{sec:ConfidenceIntervals}. MVR algorithm and its analysis are given in~\cref{Sec:SimpleR}. The experiments are presented in~\cref{Sec:Exp}. We conclude with a discussion in~\cref{Sec:Disc}.

\section{Problem Formulation and Preliminaries}
\label{sec:PF}

Consider an objective function $f: \Xc \rightarrow \Rr$, where $\Xc\subseteq \Rr^d$ is a convex and compact domain. Consider an optimal point  $x^*\in\argmax_{x\in \Xc}f(x)$. 
A learning algorithm $\Ac$ sequentially selects observation points $\{x_n\in \Xc\}_{n\in \Nn}$ and observes the corresponding noise disturbed objective values $\{y_n = f(x_n)+\epsilon_n\}_{n\in \Nn}$, where $\epsilon_n$ is the observation noise. We use the notations 
$\Hc_n=\{X_n,Y_n\}$, $X_n = [x_1,x_2,...,x_n]^{\TP}$, $Y_n = [y_1,y_2,...,y_n]^{\TP}$, $x_n\in\Xc$, $y_n\in \Rr$, for all $n\ge 1$. In a simple regret setting, the learning algorithm determines a sequence of mappings $\left\{ \Sc_n \right\}_{n\ge 1}$ where each mapping $\Sc_n: \Hc_n\rightarrow \Xc$ predicts a candidate maximizer $\hat{x}^*_n$. For algorithm $\Ac$, the simple regret under a budget of $N$ tries is defined as 
\begin{eqnarray}
r^{\Ac}_N = f(x^*) - f(\hat{x}^*_N).
\end{eqnarray}
The budget $N$ may be unknown \textit{a priori}. 
Notationwise, we use $F_n = [f(x_1),f(x_2),\dots,f(x_n)]^{\TP}$ and $E_n = [\epsilon_1,\epsilon_2,\dots,\epsilon_n]^{\TP}$ to denote the noise free part of the observations and the noise history, respectively, similar to $X_n$ and $Y_n$.

% We drop the specification of $\pi$ and $\Sc$ from the notation of observation points $\{x_n\}_{n=1}^\infty$ and the candidate maximizer $\hat{x}^*_N$, to keep the notation uncluttered. The learning algorithm shall be clear from the context throughout the paper. We also use the notations $F_n = [f(x_1),f(x_2),\dots,f(x_n)]^{\TP}$ and $E_n = [\epsilon_1,\epsilon_2,\dots,\epsilon_n]^{\TP}$ similar to $Y_n$ and $X_n$.

\subsection{Gaussian Processes}\label{Sec:GPs}

The Bayesian optimization algorithms build on GP (surrogate) models. 
A GP is a random process $\{{\hat{f}}(x)\}_{x \in \Xc}$, where each of its finite subsets follow a multivariate Gaussian distribution. The distribution of a GP is fully specified by its mean function 
$\mu(x)=\E[\hat{f}(x)]$ and a positive definite kernel (or covariance 
function) 
$k(x,x') = \E\left[(\hat{f}(x)-\mu(x))(\hat{f}(x')-\mu(x'))\right]$. Without loss of generality, it is typically assumed that $\forall x\in\Xc,\mu(x)=0$ for prior GP distributions.

Conditioning GPs on available observations provides us with powerful non-parametric Bayesian (surrogate) models over the space of functions. In particular, using the conjugate property, conditioned on $\Hc_{n}$, the posterior of $\hat{f}$ is a GP with mean function 
$\mu_{n}(x)= \E[\hat{f}(x)|\Hc_{n}] $ and kernel function 
$k_n(x,x') = \E[(\hat{f}(x)-\mu_n(x))(\hat{f}(x')-\mu_n(x'))|\Hc_{n}]$ specified as follows:
\begin{eqnarray}\nn
\mu_n(x) &=& k^{\TP}(x,X_n)\left(k(X_n,X_n)+\lambda^2 I_n\right)^{-1}Y_n,\\\label{GPn}
k_n(x,x) &=& k(x,x)- k^{\TP}(x,X_n)\left(k(X_n,X_n)+\lambda^2 I_n\right)^{-1}k(x,X_n),~
\sigma_n^2(x) =  k_n(x,x),
\end{eqnarray}
where with some abuse of notation $k(x,X_n) = [k(x,x_1),k(x,x_2),\dots,k(x,x_n)]^{\TP}$, $k(X_n,X_n)$ is the covariance matrix, $k(X_n,X_n)=[k(x_i,x_j)]_{i,j=1}^n$, $I_n$ is the identity matrix of dimension $n$ and $\lambda>0$ is a real number.

In practice, Mat{\'e}rn and squared exponential (SE) are the most commonly used kernels for Bayesian optimization~\citep[see, e.g.,][]{Shahriari2016outofloop,Snoek2012practicalBO},
\begin{eqnarray}\nn
k_{\text{Mat{\'e}rn}}(x,x') = \frac{1}{\Gamma(\nu)2^{\nu-1}}\left(\frac{\sqrt{2\nu}\rho}{l}\right)^{\nu}B_{\nu}\left(\frac{\sqrt{2\nu}\rho}{l}\right),~~~~~
 k_{\text{SE}}(x,x') = \exp \left(-\frac{\rho^2}{2l^2} \right),
\end{eqnarray}
where $l >0$ is referred to as lengthscale, $\rho=||x-x'||_{l_2}$ is the Euclidean distance between $x$ and $x'$,  $\nu>0$ is referred to as the smoothness parameter, $\Gamma$ and $B_\nu$ are, respectively, the Gamma function and the modified Bessel function of the second kind. Variation over parameter $\nu$ creates a rich family of kernels.
%where a notable example is $\nu=\frac{1}{2}$ that is referred to as the \emph{exponential kernel}. 
The SE kernel can also be interpreted as a special case of Mat{\'e}rn family when $\nu\rightarrow\infty$.

\subsection{RKHSs and Regularity Assumptions on $f$}

Consider a positive definite kernel $k:\Xc\times\Xc\rightarrow \Rr$ with respect to a finite Borel measure (e.g., the Lebesgue measure) supported on $\Xc$. A Hilbert space $H_k$ of functions on $\Xc$ equipped with an inner product $\langle \cdot, \cdot \rangle_{H_k}$ is called an RKHS with reproducing kernel $k$ if the following is satisfied.
For all $x\in\Xc$, $k(\cdot,x)\in H_k$, and
for all $x\in\Xc$ and $f\in H_k$, $\langle f,k(\cdot,x)\rangle_{H_k} = f(x)$ (reproducing property).
A constructive definition of RKHS requires the use of Mercer theorem which provides an alternative representation for kernels as an inner product of infinite dimensional feature maps~\citep[e.g.,][Theorem 4.1]{ Kanagawa2018}, and is deferred to Appendix~\tcb{B}.
We have the following regularity assumption on the objective function $f$. 

\begin{assumption}\label{Ass1}
The objective function $f$ is assumed to live in the RKHS corresponding to a positive definite kernel $k$. In particular,
$||f||_{H_k}\le B$, for some $B>0$, where $\|f\|^2_{H_k} = \langle f,f\rangle_{H_k} $. 
\end{assumption}

For common kernels, such as Mat{\'e}rn family of kernels, members of $H_k$ can uniformly approximate any continuous function on any compact subset of the domain $\Xc$~\citep{srinivas2010gaussian}. This is a very general class of functions; more general than e.g. convex or Lipschitz. It has thus gained increasing interest in recent years. 

% It is also worth mentioning that for the well studied class of Mat{\'e}rn kernels,
% the RKHS is equivalent to a Sobolev space with parameter $\nu+\frac{d}{2}$~\citep{Teckentrup2019, Kanagawa2018}. This observation provides an intuitive interpretation for the norm of Mat{\'e}rn RKHS as proportional to the cumulative $L_2$ norm of the \emph{weak derivatives} of $f$ up to $\nu+\frac{d}{2}$ order. I.e., in the case of Mat{\'e}rn family, our assumption on the norm of $f$ translates to the existence of weak derivatives of $f$ up to $\nu+\frac{d}{2}$ order which can be understood as a versatile measure for the smoothness of $f$ controlled by $\nu$. In the case of SE kernel, our regularity assumption implies the existence of all weak derivatives of $f$. For the details on the definition of weak derivatives and Sobolev spaces see~\cite{Hunter2001Analysis}.

\subsection{Regularity Assumptions on Noise} 

We consider two different cases regarding the regularity assumption on noise. 
Let us first revisit the definition of sub-Gaussian distributions. 

\begin{definition}\label{Def:subG}
A random variable $X$ is called sub-Gaussian if its moment generating function $M(h) \triangleq \E[\exp(hX)]$ is upper bounded by that of a Gaussian random variable. 
\end{definition}

The sub-Gaussian assumption implies that $\E[X] = 0$. It also allows us to use Chernoff-Hoeffding concentration inequality~\citep{antonini2008convergence} in our analysis.

We next recall the definition of light-tailed distributions.
\begin{definition}\label{def:LT}
A random variable $X$ is called light-tailed if its moment-generating function exists, i.e., there exists $h_0>0$ such that for all $|h|\le h_0$, $M(h) <\infty$.
% \begin{eqnarray}
% M(u) <\infty
% \end{eqnarray}
\end{definition}

For a zero mean light-tailed random variable $X$, we have~\citep{chareka2006locally} 
\begin{eqnarray}\label{MGFLT}
M(h)&\le& \exp(\xi_0 h^2/2),~ \forall |h|\le h_0,
\xi_0 = \sup\{M^{(2)}(h), |h|\le h_0\},
\end{eqnarray}
where $M^{(2)}(.)$ denotes the second derivative of $M(.)$ and $h_0$ is the parameter specified in Definition~\ref{def:LT}. We observe that the upper bound in~\eqref{MGFLT} is the moment generating function of a zero mean Gaussian random variable with variance $\xi_0$. Thus, light-tailed distributions are also called locally sub-Gaussian distributions~\citep{vakili2013deterministic}. %that is a generalization of sub-Gaussian distributions. 

% \subsubsection{Light-tailed Noise}

We provide confidence intervals for GP models and regret bounds for MVR under each of the following assumptions on the noise terms. 

\begin{assumption}[Sub-Gaussian Noise]\label{Ass2} The noise terms $\epsilon_n$ are i.i.d. over $n$. In addition,
$\forall h\in \Rr, \forall n\in \Nn,
\E[e^{h\epsilon_n}]\le \exp(\frac{h^2R^2}{2}),
$ for some $R>0$.
\end{assumption}

\begin{assumption}[Light-Tailed Noise]\label{Ass3} The noise terms $\epsilon_n$ are i.i.d. zero mean random variables over $n$. In addition,
$\forall h\le h_0, \forall n\in \Nn,
\E[e^{h\epsilon_n}]\le \exp(\frac{h^2\xi_0}{2}),
$ for some $\xi_0>0$.
\end{assumption}

% The light tailed assumption results in an increased confidence interval width and simple regret in comparison to the sub-Gaussian assumption as presented in~\cref{sec:ConfidenceIntervals}.

Bayesian optimization uses GP priors for the objective function $f$ and assumes a Gaussian distribution for noise (for its conjugate property). 
% Instead, our only assumptions are that the objective function $f$ is fixed and lives in an RKHS (Assumption~\ref{Ass1}) and the distribution of noise is sub-Gaussian (Assumption~\ref{Ass2}) or light tailed (Assumption~\ref{Ass3}). We 
It is noteworthy that the use of GP models is merely for the purpose of algorithm design and does not affect our regularity assumptions on $f$ and noise. We use the notation $\hat{f}$ to distinguish the GP model from the fixed $f$. 

%The assumed variance $\lambda^2$ for noise has an effect similar to the regularization parameter in kernel ridge regression~(see~\cite{Kanagawa2018}).

\subsection{Maximal Information Gain}\label{Sec:InfoGain}

The regret bounds derived in this work are given in terms of the maximal
information gain, defined as
$
\gamma_N = \sup_{X_N\subseteq \Xc}\Ic(Y_N;\hat{f}),
$
where $\Ic(Y_N;\hat{f})$ denotes the mutual information between $Y_n$ and $\hat{f}$ ~\citep[see, e.g.,][Chapter $2$]{cover1999elements}.
In the case of a GP model, the mutual information can be given as
$
\Ic(Y_n;\hat{f}) = \frac{1}{2}\log\det\left(I_n+\frac{1}{\lambda^2}k(X_n,X_n)\right),
$ where $\det$ denotes the determinant of a square matrix. 
Note that the maximal information gain is kernel-specific and $X_N$-independent.
Upper bounds on $\gamma_N$ are derived in ~\cite{srinivas2010gaussian,Janz2020SlightImprov,vakili2020information} which are commonly used to provide explicit regret bounds. In the case of Mat{\'e}rn and SE , $\gamma_N = \Oc\left(N^{\frac{d}{2\nu+d}}(\log(N))^{\frac{2\nu}{2\nu+d}}\right)$ and $\gamma_N = \Oc\left((\log(N))^{d+1}\right)$, respectively~\citep{vakili2020information}.

\section{Confidence Intervals for Gaussian Process Models}\label{sec:ConfidenceIntervals}

The analysis of bandit problems classically builds on confidence intervals applicable to the values of the objective function~\citep[see, e.g.,][]{auer2002using, bubeck2012bandits}. The GP modelling allows us to create confidence intervals for complex functions over continuous domains. In particular, we utilize the prediction ($\mu_n$) and the uncertainty estimate ($\sigma_n$) provided by GP models in building the confidence intervals which become an important building block of our analysis in the next section. To this end, we first prove the following proposition which formulates the posterior variance of a GP model as the sum of two terms: the maximum prediction error for an RKHS element from noise free observations and the effect of noise.

\begin{proposition}\label{Prop:var}
Let $\sigma^2_n$ be the posterior variance of the surrogate GP model as defined in~\eqref{GPn}. Let $Z_n^{\TP}(x) =k^{\TP}(x,X_n)\left(k(X_n,X_n)+\lambda^2 I_n\right)^{-1}$. 
%and $\zeta_i(x)=[Z_n(x)]_i$. 
We have
\begin{eqnarray}\nn
\sigma_n^2(x) = \sup_{f:||f||_{H_k}\le 1}(f(x)-Z_n^{\TP}(x)F_n)^2 + \lambda^2\|Z_n(x)\|_{l^2}^2.
\end{eqnarray}
\end{proposition}

Notice that the first term $f(x) -Z_n^{\TP}(x)F_n$ captures the maximum prediction error from noise free observations $F_n$. The second term captures the effect of noise in the surrogate GP model (and is independent of $F_n$). A detailed proof for Proposition~\ref{Prop:var} is provided in Appendix~\textcolor{blue}{C}.

Proposition~\ref{Prop:var} elicits new connections between GP models and kernel ridge regression. While the equivalence of the posterior mean in GP models and the regressor in kernel ridge regression is well known, the interpretation of posterior variance of GP models as the maximum prediction error for an RKHS element is less studied~\citep[see][Section 3, for a detailed discussion on the connections between GP models and kernel ridge regression]{Kanagawa2018}.

% We start with establishing the confidence intervals under sub-Gaussian noise. We then extend the result to the more general class of light-tailed noise. 

\subsection{Confidence Intervals under Sub-Gaussian Noise}

The following theorem provides a confidence interval for GP models applicable to RKHS elements under the assumption that the noise terms are sub-Gaussian.

\begin{theorem}\label{The:ConIneqSubG}
Assume Assumptions~\ref{Ass1} and~\ref{Ass2} hold.
Provided $n$ noisy observations $\Hc_n=\{X_n, Y_n\}$ from $f$, let $\mu_n$ and $\sigma_n$ be as defined in~\eqref{GPn}. Assume $X_n$ are independent of $E_n$. For a fixed $x\in \Xc$, define the upper and lower confidence bounds, respectively, 
\begin{eqnarray}\label{UpLow}
 U_n^{\delta}(x) \triangleq \mu_n(x) + \left(B+\beta(\delta)\right)\sigma_n(x),~
\text{and}~~~ 
 L_n^{\delta}(x) &\triangleq& \mu_n(x) - \left(B+\beta(\delta)\right)\sigma_n(x),
\end{eqnarray}
with $\beta(\delta) =  \frac{R}{\lambda} \sqrt{2\log(\frac{1}{\delta})}$, where $\delta \in (0,1)$, and $B$ and $R$ are the parameters specified in Assumptions~\ref{Ass1} and~\ref{Ass2}. We have
\begin{eqnarray}\nn
f(x)  \le U_n^{\delta}(x)~~~\text{w.p. at least}~1-\delta,~
\text{and}~~~
f(x)  &\ge& L_n^{\delta}(x)~~~\text{w.p. at least}~1-\delta.
\end{eqnarray}
\end{theorem}

We can write the difference in the objective function and the posterior mean as follows.
\begin{eqnarray}\nn
f(x) - \mu_n(x) = f(x) - Z_n^{\TP}(x)Y_n = \underbrace{
f(x) - Z_n^{\TP}(x)F_n}_{\text{Prediction error from noise free observations}} -\underbrace{Z_n^{\TP}(x)E_n}_{\text{The effect of noise}}.
\end{eqnarray}
%Both these terms are then bounded by a factor of posterior standard deviation based on Proposition~\ref{Prop:var}. 
The first term can be bounded directly following Proposition~\ref{Prop:var}. The second term is bounded as a result of Proposition~\ref{Prop:var} and Chernoff-Hoeffding inequality. A detailed proof of Theorem~\ref{The:ConIneqSubG} is provided in Appendix~\textcolor{blue}{D}.

\subsection{Confidence Intervals under Light-Tailed Noise}

% We have the following extended Chernoff-Hoeffding bound 
% \begin{lemma}[Chernoff-Hoeffding Bound for Light-Tailed Distributions]\label{Lemma:LT} Let $X$ be a random variable with a ailed distribution. We have for all $\delta\in[0,\xi h_0]$, $a\in (0, \frac{1}{2\xi}]$,
% \begin{eqnarray}
% \Pr[|X-\mu|\ge \delta]\le 2\exp(-a\delta^2)
% \end{eqnarray}

% \end{lemma}

% Proven in~\cite{}, Lemma~\ref{Lemma:LT} extends the Chernoff-Hoeffding bound given in Lemma~\ref{Lemma:SG} to light-tailed distributions. Building on this lemma and following similar steps as in the proof of Theorem~\ref{The:SG} we prove the following theorem. 

We now extend the confidence intervals to the case of light-tailed noise. The main difference with sub-Gaussian noise is that Chernoff-Hoeffding inequality is no more applicable. We derive new bounds accounting for light-tailed noise in the analysis of Theorem~\ref{The:ConIneqLT}.

\begin{theorem}\label{The:ConIneqLT}
Assume Assumptions~\ref{Ass1} and~\ref{Ass3} hold. For a fixed $x\in\Xc$, define the upper and lower confidence bounds $U_n^{\delta}(x)$ and $L_n^{\delta}(x)$ similar to Theorem~\ref{The:ConIneqSubG} with $\beta(\delta) =  \frac{1}{\lambda} \sqrt{2\left(\xi_0\vee\frac{2\log(\frac{1}{\delta})}{h_0^2}\right)\log(\frac{1}{\delta})}$~\footnote{The notation $\vee$ is used to denote the maximum of two real numbers, $\forall a,b\in \Rr, (a\vee b)\triangleq \max(a,b)$.}, where $\delta\in(0,1)$, and $B$, $h_0$ and $\xi_0$ are specified in Assumptions~\ref{Ass1} and~\ref{Ass3}. Assume $X_n$ are independent of $E_n$. We have
\begin{eqnarray}\nn
f(x)  \le U_n^{\delta}(x)~~~\text{w.p. at least}~1-\delta,~
\text{and}~~~
f(x)  &\ge& L_n^{\delta}(x)~~~\text{w.p. at least}~1-\delta.
\end{eqnarray}

% \begin{eqnarray}\nn
% f(x)  &\le& U_n^{\delta}(x)~~~\text{w.p.}~1-\delta\\
% f(x)  &\ge& L_n^{\delta}(x)~~~\text{w.p.}~1-\delta.
% \end{eqnarray}
% \begin{eqnarray}\nn
% f(x)  &\le& \mu_n(x) +\sigma_n(x)\left( B+ \frac{1}{\lambda} \sqrt{2\left(\xi_0\vee\frac{2\log(\frac{1}{\delta})}{h_0^2})\log(\frac{1}{\delta}\right)}\right)~~~\text{w.p.}~1-\delta\\\nn
% f(x)  &\ge& \mu_n(x)-\sigma_n(x)\left( B+ \frac{1}{\lambda} \sqrt{2\left(\xi_0\vee\frac{2\log(\frac{1}{\delta})}{h_0^2})\log(\frac{1}{\delta}\right)}\right)~~~\text{w.p.}~1-\delta.
% \end{eqnarray}
\end{theorem}

In comparison to Theorem~\ref{The:ConIneqSubG}, under the light-tailed assumption, the confidence interval width increases with a multiplicative $\Oc(\sqrt{\log(\frac{1}{\delta})})$ factor. A detailed proof of Theorem~\ref{The:ConIneqLT} is provided in Appendix~\textcolor{blue}{D}.

\begin{remark}
Theorems~\ref{The:ConIneqSubG} and~\ref{The:ConIneqLT} rely on the assumption that $X_n$ are independent of $E_n$. As we shall see in~\cref{Sec:SimpleR}, this assumption is satisfied when the confidence intervals are applied to the analysis of MVR.
\end{remark}
% \subsection{Heavy-Tailed Noise}

%%%

\subsection{Comparison with the Existing Confidence Intervals}\label{comp:conf}

The most relevant work to our Theorems~\ref{The:ConIneqSubG} and~\ref{The:ConIneqLT} is~\citep[][Theorem $2$]{Chowdhury2017bandit} which itself was an improvement over~\citep[][Theorem $6$]{srinivas2010gaussian}.  \cite{Chowdhury2017bandit} built on feature space representation of GP kernels and self-normalized martingale inequalities~\citep{Abbasi2011, pena2008self} to establish a $1-\delta$ confidence interval in the same form as in Theorem~\ref{The:ConIneqSubG}, under Assumptions~\ref{Ass1} and~\ref{Ass2}, with confidence interval width 
$B+R\sqrt{2(\gamma_n+1+\log(\frac{1}{\delta}))}
$~\footnote{The effect of $\lambda$ is absorbed in $\gamma_n$.} (instead of $B+\beta(\delta)$). 
There is a stark contrast between this confidence interval and the one given in Theorem~\ref{The:ConIneqSubG} in its dependence on $\gamma_n$ which has a relatively large and possibly polynomial in $n$ value. That contributes an extra $\Oc(\sqrt{\gamma_N})$ multiplicative factor to regret.

Neither of these two results (our Theorem~\ref{The:ConIneqSubG} and~\citep[][Theorem~$2$]{Chowdhury2017bandit}) imply the other. Although our confidence interval is much tighter, there are two important differences in the settings of these theorems. One difference is in the probabilistic dependencies between the observation points $x_n$ and the noise terms $\{\epsilon_j\}_{j<n}$. While Theorem~\ref{The:ConIneqSubG} assumes that $X_n$ are independents of $E_n$,~\citep[][Theorem~$2$]{Chowdhury2017bandit} allows for the dependence of $x_n$ on the previous noise terms~$\{\epsilon_j\}_{j<n}$. This is a reflection of the difference in the analytical requirements of MVR and GP-UCB. The other difference is that~\citep[][Theorem~$2$]{Chowdhury2017bandit} holds for all $x\in \Xc$. While, Theorem~\ref{The:ConIneqSubG} holds for a single $x\in \Xc$. As we will see in~\cref{Sec:analysis}, a probability union bound can be used to obtain confidence intervals applicable to all $x$ in (a discretization of) $\Xc$, which contributes only logarithmic terms to regret in contrast to $\Oc(\sqrt{\gamma_n})$. Roughly speaking, we are trading off the extra $\Oc(\sqrt{\gamma_n})$ term for restricting the confidence interval to hold for a single $x$. It remains an open problem whether the same can be done when $x_n$ are allowed to depend on $\{\epsilon_j\}_{j<n}$.

\section{Maximum Variance Reduction and Simple Regret}\label{Sec:SimpleR}

In this section, we first formally present an exploration policy based on GP models referred to as Maximum Variance Reduction (MVR). We then utilize the confidence intervals for GP models derived in~\cref{sec:ConfidenceIntervals} to prove bounds on the simple regret of MVR. 

\subsection{Maximum Variance Reduction Algorithm}
MVR relies on the principle of reducing the maximum uncertainty where the uncertainty is measured by the posterior variance of the GP model. After $N$ exploration trials, MVR returns a candidate maximizer according to the prediction provided by the learnt GP model. A pseudo-code is given in Algorithm~\ref{Alg1}.

% \begin{algorithm}[tb]
%   \caption{Maximum Variance Reduction (MVR)}
%   \label{Alg1}
% \begin{algorithmic}
%   \STATE {\bfseries Initialization:} $k$, $\Xc$, $f$, $\sigma_0(x)\equiv 1$.
%   %\REPEAT
%   %\STATE Initialize $noChange = true$.
%   \FOR{$n=1,2,\dots, N$}
%   %{\bfseries to} $m-1$}
%   \STATE $x_n = \argmax_{x\in\Xc}\sigma_{n-1}(x)$ (tie is broken arbitrarily).
%   \STATE Update $\sigma_{n}(x)$ according to~\eqref{GPn}.
%   \ENDFOR
%   \STATE Update $\mu_{N}(x)$ according to~\eqref{GPn}.
%   \STATE \textbf{return} $\hat{x}^*_N = \argmax_{x\in\Xc}\mu_{N}(x)$ (tie is broken arbitrarily).
%   %\UNTIL{$noChange$ is $true$}
% \end{algorithmic}
% \end{algorithm}

\begin{algorithm}
\caption{Maximum Variance Reduction (MVR)}\label{Alg1}
\begin{algorithmic}[1]

\State \textbf{Initialization:  $k$, $\Xc$, $f$, $\sigma^2_0(x)= k(x,x)$.} 
\For{$n=1,2,\dots,N$}
\State $x_n = \argmax_{x\in\Xc}\sigma^2_{n-1}(x)$, where a tie is broken arbitrarily.
\State Update $\sigma^2_{n}(.)$ according to~\eqref{GPn}.
\EndFor\label{euclidendwhile}
\State Update $\mu_{N}(.)$ according to~\eqref{GPn}
\State \textbf{return} $\hat{x}^*_N = \argmax_{x\in\Xc}\mu_{N}(x)$, where a tie is broken arbitrarily.
\end{algorithmic}
\end{algorithm}

\subsection{Regret Analysis}\label{Sec:analysis}

For the analysis of MVR, we assume there exists a fine discretization of the domain for RKHS elements, which is a standard assumption in the literature~\citep[see, e.g.,][]{srinivas2010gaussian, Chowdhury2017bandit, Vakili2020Scalable}.

\begin{assumption}\label{ass4}
For each given $n\in\Nn$ and $f\in H_k$ with $\|f\|_{H_k}\le B$, there exists a discretization $\Dc_n$ of $\Xc$ such that $f(x) - f([x]_n)\le \frac{1}{\sqrt{n}}$, where $[x]_n = \argmin_{x'\in \Dc_n}||x'-x||_{l^2}$ is the closest point in $\Dc_n$ to $x$, and $|\Dc_n|\le CB^dn^{d/2}$, where $C$ is a constant independent of $n$ and $B$.
\end{assumption}
Assumption~\ref{ass4} is a mild assumption that holds for typical kernels such as SE and Mat{\'e}rn~\citep{srinivas2010gaussian, Chowdhury2017bandit}.
The following theorem provides a high probability bound on the regret performance of MVR when the noise terms satisfy either Assumption~\ref{Ass2} or~\ref{Ass3}. 

\begin{theorem}\label{The:regSubG}
Consider the Gaussian process bandit problem. Under Assumptions~\ref{Ass1},~\ref{ass4}, and~(\ref{Ass2} or~\ref{Ass3}), for $\delta\in(0,1)$, with probability at least $1-\delta$, MVR satisfies
\begin{eqnarray}\nn
r^{\text{MVR}}_N &\le& \sqrt{\frac{2\gamma_N}{\log(1+\frac{1}{\lambda^2})N}}
\left(
2B
+\beta(\frac{\delta}{3}) 
+ \beta\bigg(\frac{\delta}{
3C\left(B+\sqrt{N}\beta(2\delta/3N)\right)^dN^{d/2}
}\bigg)
\right) 
+ \frac{2}{\sqrt{N}},
\end{eqnarray}
where under Assumption~\ref{Ass2}, $\beta(\delta) =  \frac{R}{\lambda} \sqrt{2\log(\frac{1}{\delta})}$, and under Assumption~\ref{Ass3}, $\beta(\delta) = \frac{1}{\lambda} \sqrt{2\left(\xi_0\vee\frac{2\log(\frac{1}{\delta})}{h_0^2}\right)\log(\frac{1}{\delta})}$,
and
$B$, $R$, $h_0$, $\xi_0$, and $C$ are the constants specified in Assumptions~\ref{Ass1},~\ref{Ass2},~\ref{Ass3} and~\ref{ass4}.
% where $c_1 = \frac{2}{\log(1+\frac{1}{\lambda^2})}$
\end{theorem}

% We also have the following theorem on the regret performance of MVR when the noise terms are light-tailed.
% \begin{theorem}\label{The:regLT}
% Consider the Gaussian process bandit problem. Under Assumptions~\ref{Ass1} and~\ref{Ass3}, with probability $1-\delta$, MVR satisfies

% \begin{eqnarray}\nn
% r^{\text{MVR}}_N \le&& \\\nn
% &&\hspace{-5em}\sqrt{\frac{2\gamma_N}{\log(1+\frac{1}{\lambda^2})N}}
% \left( B+ \frac{1}{\lambda} \sqrt{2\left(\xi_0\vee\frac{2\log(\frac{1}{\delta})}{h_0^2})\log(\frac{1}{\delta}\right)}\right),
% \end{eqnarray}
% where $B$, $\zeta_0$ and $h_0$ are the parameters specified in Assumptions~\ref{Ass1} and~\ref{Ass3}.
% % where $c_1 = \frac{2}{\log(1+\frac{1}{\lambda^2})}$
% \end{theorem}

% High probability bounds on regret are stronger than bounds on average regret. Since the objective function is bounded, selecting $\delta =\frac{1}{N}$ leads to $\E[r^{\text{MVR}}_N] = O\left(\sqrt{\frac{\gamma_N\log(N)}{N}}\right)$ and $\E[r^{\text{MVR}}_N] = O\left(\sqrt{\frac{\gamma_N}{N}}\log(N)\right)$, respectively, under assumptions of Theorem~\ref{The:regSubG} and Theorem~\ref{The:regLT}. 
A detailed proof of the theorem is provided in Appendix~\textcolor{blue}{E}.
\begin{remark}
Under Assumptions~\ref{Ass2} and~\ref{Ass3}, respectively, the regret bounds can be simplified as
\begin{eqnarray}\nn
r^{\text{MVR}}_N = {\Oc}(\sqrt{\frac{\gamma_N\log(N^{d}/\delta)}{N}}),~
\text{and}~~~
r^{\text{MVR}}_N = {\Oc}\left(\sqrt{\frac{\gamma_N}{N}}\log(N^{d}/\delta)\right).
\end{eqnarray}
% $r^{\text{MVR}}_N = {\Oc}(\sqrt{\frac{\gamma_N\log(N^{d}/\delta)}{N}})$ and $r^{\text{MVR}}_N = {\Oc}\left(\sqrt{\frac{\gamma_N}{N}}\log(N^{d}/\delta)\right)$ 
For instance, in the case of Mat{\'e}rn-$\nu$ kernel, under Assumption~\ref{Ass2} and~\ref{Ass3}, respectively,
\begin{eqnarray}\nn
r^{\text{MVR}}_N =\Oc\left(N^{\frac{-\nu}{2\nu+d}}(\log(N))^{\frac{\nu}{2\nu+d}}\sqrt{\log(N^{d}/\delta)} \right),~
\text{and}~
r^{\text{MVR}}_N =\Oc\left(N^{\frac{-\nu}{2\nu+d}}(\log(N))^{\frac{\nu}{2\nu+d}}\log(N^{d}/\delta) \right),
\end{eqnarray}
which always converge to zero as $N$ grows (unlike the existing regret bounds). 
\end{remark}

\begin{remark}
In the analysis of Theorem~\ref{The:regSubG}, we apply Assumption~\ref{ass4} to $\mu_N$ as well as $f$. For this purpose, we derive a high probability $B+\sqrt{N}\beta(2\delta/3N)$ upper bound on $\|\mu_N\|_{H_k}$ (see Lemma~\tcb{4} in Appendix~\tcb{E}), which appears in the regret bound expression. 
\end{remark}

\subsection{Optimal Order Simple Regret with SE and Mat{\'e}rn Kernels}
To enable a direct comparison with the lower bounds on simple regret proven in~\cite{Scarlett2017Lower, cai2020lower}, in the following corollary, we state a dual form of Theorem~\ref{The:regSubG} for the Mat{\'e}rn and SE kernels. Specifically we formalize the number of exploration trials required to achieve an average $\epsilon$ regret.

\begin{corollary}\label{cor1}
Consider the GP bandit problem with an SE or a Mat{\'e}rn kernel. For $\epsilon\in(0,1)$, define
$
N_{\epsilon}=\min\{N\in \Nn: \E[{r}^{\text{MVR}}_n]\le \epsilon, \forall n\ge N\}.
$
Under Assumptions~\ref{Ass1},~\ref{ass4}, and~(\ref{Ass2} or~\ref{Ass3}), upper bounds on $N_\epsilon$ are reported in Table~\ref{Table:cor1}.
\begin{table}[H]
  \caption{The upper bounds on $N_\epsilon$ defined in~Corollary~\ref{cor1} with SE or Mat{\'e}rn kernel.}
  \label{Table:cor1}
  \centering
  \begin{tabular}{c c c}
    \toprule
    Kernel& Under Assumption~\ref{Ass2} &
    Under Assumption~\ref{Ass3}  \\
    \midrule
    SE&$N_{\epsilon} = \Oc\left((\frac{1}{\epsilon})^2\log(\frac{1}{\epsilon})^{d+2}  
    \right)
    $&$N_{\epsilon} = \Oc\left((\frac{1}{\epsilon})^2\log(\frac{1}{\epsilon})^{d+3}  
    \right)
    $\\
    %%%
    Mat{\'e}rn-$\nu$ & $N_{\epsilon} = \Oc\left((\frac{1}{\epsilon})^{2+\frac{d}{\nu}}
    (\log(\frac{1}{\epsilon})^{\frac{4\nu+d}{2\nu}})\right)
    $&$
    N_{\epsilon} = \Oc\left((\frac{1}{\epsilon})^{2+\frac{d}{\nu}}
    (\log(\frac{1}{\epsilon})^{\frac{6\nu+2d}{2\nu}})\right)
    $\\
    \bottomrule
  \end{tabular}
\end{table}
% In the case of SE kernel:
% Under Assumptions~\ref{Ass1} and~\ref{Ass2}, we have
% %\begin{eqnarray}
% $N = \Oc\left((\frac{1}{\epsilon})^2\log(\frac{1}{\epsilon})^{d+2}  
% \right).
% $%\end{eqnarray} 
% Under Assumptions~\ref{Ass1} and~\ref{Ass3}, we have
% %\begin{eqnarray}\nn
% $N = \Oc\left((\frac{1}{\epsilon})^2\log(\frac{1}{\epsilon})^{d+3}  
% \right).
% $%\end{eqnarray}
% In the case of Mat{\'e}rn kernel:
% Under Assumptions~\ref{Ass1} and~\ref{Ass2}, we have
% %\begin{eqnarray}
% $N = \Oc\left((\frac{1}{\epsilon})^{2+\frac{d}{\nu}}
% (\log(\frac{1}{\epsilon})^{\frac{4\nu+d}{2\nu}})\right).
% $%\end{eqnarray}
% Under Assumptions~\ref{Ass1} and~\ref{Ass3}, we have
% %\begin{eqnarray}
% $
% N = \Oc\left((\frac{1}{\epsilon})^{2+\frac{d}{\nu}}
% (\log(\frac{1}{\epsilon})^{\frac{6\nu+2d}{2\nu}})\right).
% $%\end
\end{corollary}

A proof is provided in Appendix~\textcolor{blue}{F}.
\cite{Scarlett2017Lower, cai2020lower}
showed that for the SE kernel, an average simple regret of $\epsilon$ requires $N_{\epsilon}= \Omega\left(\frac{1}{\epsilon^2}(\log(\frac{1}{\epsilon}))^{\frac{d}{2}}\right)$. For the Mat{\'e}rn-$\nu$ kernel they gave the analogous bound of $N_{\epsilon}=\Omega\left((\frac{1}{\epsilon})^{2+\frac{d}{\nu}}\right)$. They also reported significant gaps between these lower bounds and the existing results~\citep[see, e.g.,][Table~I]{Scarlett2017Lower}. Comparing with Corollary~\ref{cor1}, our bounds are tight in all cases up to $\log(1/\epsilon)$ factors.

\section{Experiments}\label{Sec:Exp}

In this section, we provide numerical experiments on the simple regret performance of MVR, Improved GP-UCB (IGP-UCB) as presented in~\cite{Chowdhury2017bandit}, and GP-PI and GP-EI as presented in~\cite{hoffman2011portfolio}.

\begin{figure}[H]
\centering
    %%%
    \subfloat[SE, Gaussian Noise]{\label{fig:branin_plot}\centering \includegraphics[width=2.4in, height =1.0in]{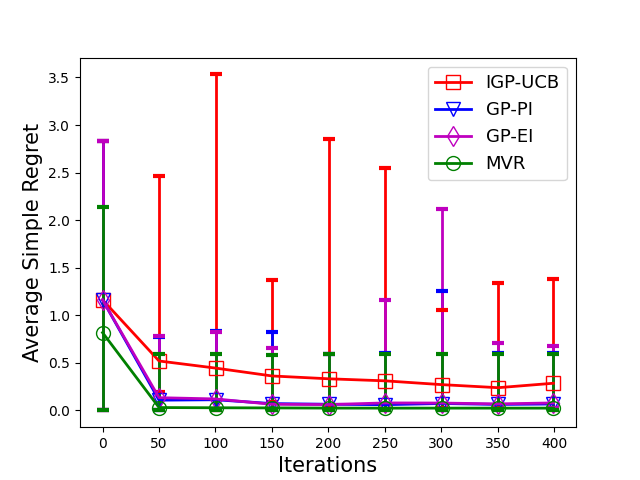}}
    %%%
    \subfloat[Mat{\'e}rn, Gaussian Noise]{\label{fig:rosenbrock_plot}\centering \includegraphics[width=2.4in, height =1.0in]{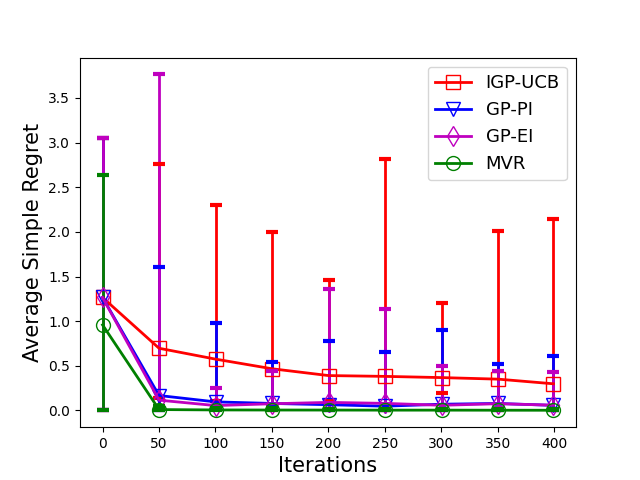}}
    
    %%%
    \subfloat[SE, Laplace Noise]{\label{fig:cnn_plot}\centering \includegraphics[width=2.4in, height =1.0in]{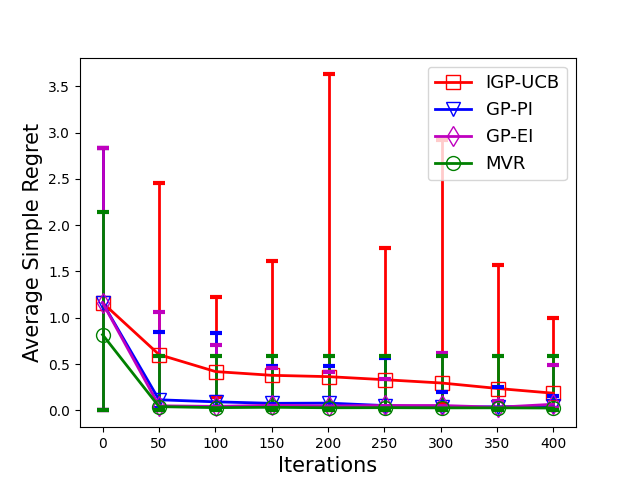}}
    %%%
    \subfloat[Mat{\'e}rn, Laplace Noise]{\label{fig:time_taken} \centering \includegraphics[width=2.4in, height =1.0in]{Figures/RKHS_SE_laplace_error_bars.png}}

\caption{Comparison of the simple regret performance of Bayesian optimization algorithms on samples from RKHS.}
\label{Fig1}
\end{figure}

We follow the experiment set up in \cite{Chowdhury2017bandit} to generate test functions from the RKHS. First, $100$ points are uniformly sampled from interval $[0,1]$. A GP sample with kernel $k$ is drawn over these points. Given this sample, the mean of posterior distribution is used as the test function $f$. Parameter $\lambda^2$ is set to $1\%$ of the function range. For IGP-UCB we set the parameters exactly as described in~\cite{Chowdhury2017bandit}. The GP model is equipped with SE or Mat{\'e}rn-$2.5$ kernel with $l=0.2$. We use $2$ different models for the noise: a zero mean Gaussian with variance equal to $\lambda^2$ (a sub-Gaussian distribution) and a zero mean Laplace with scale parameter equal to $\lambda$ (a light-tailed distribution). 
We run each experiment over 25 independent trials and plot the average simple regret in Figure~\ref{Fig1}. %The regret curves have been smoothed out using Savitzky-Golay filters.
More experiments on two commonly used benchmark functions for Bayesian optimization (Rosenbrock and Hartman$3$) are reported in Appendix~\tcb{G}. Further details on the experiments 
%and the source code 
are provided in the supplementary material.

\section{Discussion}\label{Sec:Disc}

In this paper, we proved novel and sharp confidence intervals for GP models applicable to RKHS elements. 
% To this end,
% we formulated the posterior variance of a GP
% model as the sum of two terms: the maximum prediction
% error for an RKHS element from noise-free observations, and the effect of noise. 
We then built on these results to prove $\tilde{\Oc}(\sqrt{\gamma_N/N})$ bounds for the simple regret of an adaptive exploration algorithm under the framework of GP bandits. In addition, for the practically relevant SE and Mat{\'e}rn kernels, where a lower bound on regret is known~\cite{Scarlett2017Lower, cai2020lower}, we showed the order optimality of our results up to logarithmic factors. That closes a significant gap in the literature of analysis of Bayesian optimization algorithms under the performance measure of simple regret.

The limitation of our work adhering to simple regret is that neither our theoretical nor experimental result proves that MVR is a better algorithm in practice. Overall, exploration-exploitation oriented algorithms such as GP-UCB may perform worse than MVR in terms of simple regret due to two reasons. One is over-exploitation of local maxima when $f$ is multi-modal, and the other is dependence on an exploration-exploitation balancing hyper-parameter that is often set too conservatively, to guarantee low regret bounds. Furthermore, their existing analytical regret bounds are suboptimal and possibly vacuous (non-diminishing; when $\gamma_N$ grows faster than $\sqrt{N}$, as discussed). On the other hand, when compared in terms of \emph{cumulative} regret ($\sum_{n=1}^Nf(x^*)-f(x_n)$), MVR suffers from a linear regret. 

The main value of our work is in proving tight bounds on the simple regret of a GP based exploration algorithm, when other Bayesian optimization algorithms such as GP-UCB lack a proof for an always diminishing and non-vacuous regret under the same setting as ours. It remains an open question whether the possibly vacuous regret bounds of GP-UCB (as well as GP-TS and GP-EI whose analysis is inspired by that of GP-UCB) is a fundamental limitation or an artifact of its proof. %Our emphasis on GP-UCB is due to the fact that the analysis of other algorithms (e.g. GP-TS, GP-EI) are inspired by the its analysis.   

It is worth reiterating that simple regret is favorable in situations with a preliminary exploration phase (for instance hyper-parameter tuning)~\citep{bubeck2011pure}. It has been explicitly studied under numerous settings, e.g.,
~\citep[][Lipschitz continuous $f$]{bubeck2011pure, carpentier2015simple, deshmukh2018simple},~\citep[][$f$ in RKHS, noise-free observations]{Bull2011EI},~\citep[][a known prior distribution on $f$, noise-free observations]{grunewalder2010regret,Freitas2012, Kawaguchi2015}, \citep[][a known prior distribution on $f$, noisy observations]{contal2013parallel},
\citep[][$f$ in RKHS, noisy observations]{Scarlett2017Lower, cai2020lower,shekhar2020multi, bogunovic2016truncated}. See also~\cref{sec:LitRef} and Appendix~\tcb{A} for comparison with existing results including~\cite{shekhar2020multi, bogunovic2016truncated}.

% \begin{ack}
% Use unnumbered first level headings for the acknowledgments. All acknowledgments
% More information about this disclosure can be found at: \url{https://neurips.cc/Conferences/2021/PaperInformation/FundingDisclosure}.

% Do {\bf not} include this section in the anonymized submission, only in the final paper. You can use the \texttt{ack} environment provided in the style file to autmoatically hide this section in the anonymized submission.
% \end{ack}

\medskip

\newpage

\bibliography{references}
\bibliographystyle{abbrvnat}

\medskip
\vspace{2em}

\begin{appendix}

\section{Further Comparison with the Existing Regret Bounds}~\label{app:compLR}

% \subsection{Comparison with the Existing Regret Bounds}\label{comp:reg}

There are several Bayesian optimization algorithms namely GP-UCB~\citep{srinivas2010gaussian}, IGP-UCB, GP-TS~\citep{Chowdhury2017bandit}, TruVar~\citep{bogunovic2016truncated}, GP-EI~\citep{wang2014theoreticalGPEI,nguyen2017regretGPEI} and KernelUCB~\citep{Valko2013kernelbandit} which enjoy theoretical upper bounds on regret (under Assumptions~\tcb{$1$},~\tcb{$2$} and~\tcb{$4$}), which grow at least as fast as $\Oc(\frac{\gamma_N}{\sqrt{N}})$.
These bounds do not necessarily converge to zero, since $\gamma_N$ can grow faster than $\sqrt{N}$ resulting in vacuous regret bounds. For example, in the case of a Mat{\'e}rn-$\nu$ kernel, replacing $\gamma_N = \tilde{\Oc}(N^{\frac{d}{2\nu+d}})$~\citep{vakili2020information} results in an $\tilde{\Oc}(N^{\frac{d-2\nu}{4\nu+2d}})$ regret which does not converge to zero for $d>2\nu$, meaning the algorithm does not necessarily approach $f(x^*)$. \cite{Janz2020SlightImprov}
developed a GP-UCB based
algorithm, specific to Mat{\'e}rn family of kernels, that constructs a cover for the search space, as many hypercubes, and fits an independent GP to each cover element. This algorithm, referred to as $\pi$-GP-UCB,
was proven to achieve diminishing regret for all $\nu>1$ and $d\ge 1$. 
Recently,~\cite{shekhar2020multi} introduced LP-GP-UCB where the GP model is augmented with local polynomial estimators to construct a multi-scale upper confidence bound guiding the sequential optimization. They further improved the regret bounds of~\cite{Janz2020SlightImprov} and showed that LP-GP-UCB matches the lower bounds for some configuration of parameters $\nu$ and $d$ in the case of a Mat{\'e}rn kernel. Defining $\bm{I}_0 = (0,1]$, $\bm{I}_1 = (1,\frac{d(d+1)}{2}]$, $\bm{I}_2 = (\frac{d(d+1)}{2},\frac{d^2+5d+12}{4}]$ and $\bm{I}_3 = (0,\infty) \setminus \bm{I}_0\cup\bm{I}_1\cup\bm{I}_2 $, their bounds on simple regret are as follows. For $\nu\in \bm{I}_0\cup\bm{I}_1$, $r_N^{\text{LP-GP-UCB}} = \tilde{\Oc}(N^{\frac{-\nu}{2\nu+d}})$. For $\nu\in \bm{I}_2$, $r_N^{\text{LP-GP-UCB}} = \tilde{\Oc}(N^{\frac{-1}{2+d}})$. For $\nu\in \bm{I}_3$, $r_N^{\text{LP-GP-UCB}} = \tilde{\Oc}(N^{\frac{-4\nu+d(d+1)}{8\nu+2d(d+5)}})$~\cite[see,][Sec. $3.2$, for a detailed discussion on the bounds on the simple regret of LP-GP-UCB]{shekhar2020multi}.
%~\citep[see also][Appendix A.4, for a discussion]{cai2020lower}. 
In comparison, our bounds on simple regret match the $\Omega(N^{\frac{-\nu}{2\nu+d}})$ lower bound, up to logarithmic factors, with all parameters $\nu$ and $d$. In addition, LP-GP-UCB is impractical due to large constant factors, though a practical heuristic was also given. While, MVR enjoys a simple implementation and works efficiently in practice.
Of important theoretical value, SupKernelUCB~\cite{Valko2013kernelbandit}, which builds on episodic independent batches of observations was proven to achieve $\tilde{\Oc}(\sqrt{\frac{\gamma_N}{N}})$ regret on a finite set ($|\Xc|<\infty$). 
% It is discussed that the finite set assumption can be relaxed to continuous spaces~\cite{Janz2020SlightImprov, cai2020lower}. However, 
SupKernelUCB is also reported to perform poorly in practice~\citep{Janz2020SlightImprov,Calandriello2019Adaptive,  cai2020lower}.

% A comparison between these bounds and the upper bounds provided in Theorem
% ~\ref{The:regSubG} shows an $\Oc(\sqrt{\gamma_N})$ improvement in our results. That is a direct consequence of $\Oc(\sqrt{\gamma_N})$ improvement in our confidence intervals for GP models. In addition since $\gamma_N$ is sublinear in $N$, the simple regret of MVR always converges to $0$.

It is noteworthy that our techniques do not directly apply to the analysis of cumulative regret of algorithms such as GP-UCB. The key difference is that in MVR the observation points $x_n$ are independent of the noise terms $\epsilon_n$ (although $x_n$ are allowed to depend on $\{x_j\}_{j<n}$, and $\hat{x}^*_N$ is allowed to depend on $\{x_n,\epsilon_n\}_{n\le N}$), while in GP-UCB $x_n$ are allowed to depend on $\{\epsilon_j\}_{j< n}$ (see also Sec. \tcb{$3.3$}). 
It remains an interesting open question whether the state of the art upper bound on the regret performance of GP-UCB~\citep{Chowdhury2017bandit} is tight or the gap with the lower bound~\citep{Scarlett2017Lower} is an artifact of its proof.

\section{Constructive Definition of RKHS}\label{app:RKHS}

A constructive definition of RKHS requires the use of Mercer theorem which provides an alternative representation for kernels as an inner product of infinite dimensional feature maps~\citep[see, e.g.,][Theorem $4.1$]{Kanagawa2018}.

%\begin{theorem}[Mercer's Theorem]\label{The:Mercer}
\paragraph{Mercer Theorem:}
Let $k$ be a continuous kernel with respect to a finite Borel measure.
There exists $\{(\lambda_m,\phi_m)\}_{m=1}^{\infty}$ such that $\lambda_m\in \Rr^{+}$, $\phi_m\in H_k$, for $m\ge1$, and
\begin{eqnarray}\nn
k(x,x') = \sum_{m=1}^{\infty} \lambda_m\phi_m(x)\phi_m(x').
\end{eqnarray}
%\end{theorem}

%The $\{\lambda_m\}_{m=1}^\infty$ and the $\{\phi_m\}_{m=1}^\infty$ are referred to as the eigenvalues and the eigenfeatures (or eigenfunctions) of $k$, respectively. 

The RKHS can consequently be represented in terms of $\{(\lambda_m,\phi_m)\}_{m=1}^{\infty}$ using Mercer's representation theorem~\citep[see, e.g.,][ Theorem $4.2$]{ Kanagawa2018}.

%\begin{theorem}[Mercer's Representation Theorem]\label{The:MercerRep} 
\paragraph{Mercer's Representation Theorem:}
Let $\{(\lambda_m,\phi_m)\}_{m=1}^\infty$ be the same as in Mercer Theorem.
%~\ref{The:Mercer}. 
Then, the RKHS of $k$ is given by
\begin{eqnarray}\nn
\scriptsize
H_k = \left\{ f(\cdot)=\sum_{m=1}^{\infty}w_m\lambda_{m}^{\frac{1}{2}}\phi_m(\cdot): ||f||^2_{H_k} \triangleq \sum_{m=1}^\infty w_m^2<\infty \right\}.
\end{eqnarray}
%\end{theorem}
Mercer's representation theorem
indicates that $\{\lambda_m^{\frac{1}{2}}\phi_m\}_{m=1}^\infty$ form an orthonormal basis for $H_k$. It also provides a constructive definition for the RKHS as the span of this orthonormal basis, and a definition for the norm of a member $f$ as the $l_2$ norm of the weights $w_m$. 

The RKHS of Mat{\'e}rn is equivalent to a Sobolev space with parameter $\nu+\frac{d}{2}$~\citep{Kanagawa2018,Teckentrup2019}. This observation provides an intuitive interpretation for the norm of Mat{\'e}rn RKHS as proportional to the cumulative $L_2$ norm of the \emph{weak derivatives} of $f$ up to $\nu+\frac{d}{2}$ order. I.e., in the case of Mat{\'e}rn family, Assumption~\tcb{$1$} on the norm of $f$ translates to the existence of weak derivatives of $f$ up to $\nu+\frac{d}{2}$ order which can be understood as a versatile measure for the smoothness of $f$ controlled by $\nu$. In the case of SE kernel, the regularity assumption implies the existence of all weak derivatives of $f$. For the details on the definition of weak derivatives and Sobolev spaces see~\cite{Hunter2001Analysis}. 

\section{Proof of Proposition \textcolor{blue}{$1$}}\label{app:prop1}

Recall the  notations $Y_n = [y_1, y_2, \dots, y_n]^{\TP}$, $F_n = [f(x_1), f(x_2), \dots, f(x_n)]^{\TP}$, $Z_n^{\TP}(x) =k^{\TP}(x,X_n)\left(k(X_n,X_n)+\lambda^2 I_n\right)^{-1}$. Let $\zeta_i(x)=[Z_n(x)]_i$.
From the closed form expression for the posterior mean of GP models, we have $\mu_n(x) = Z^{\TP}_n(x)Y_n$.

The proof of Proposition \textcolor{blue}{$1$} uses the following lemma.

\begin{lemma}

For a positive definite kernel $k$ and its corresponding RKHS, the following holds.  
\begin{eqnarray}\label{xx1}
\sup_{f: ||f||_{H_k}\le 1}\left(f(x) - \sum_{i=1}^n \zeta_i(x)f(x_i)\right)^2 = \bigg|\bigg| k(.,x) - \sum_{i=1}^n \zeta_i(x) k(.,x_i) \bigg|\bigg|_{H_k}^2.
\end{eqnarray}
\end{lemma}

The lemma establishes the equivalence of the RKHS norm of a linear combination of feature vectors induced by $k$ to the supremum of the linear combination of the corresponding function values, over the functions in the unit ball of the RKHS. For a proof, see~\citep[][Lemma $3.9$]{Kanagawa2018}. 

% gives an expression for the RKHS norm of a linear combination of feature vectors induced by $k$ as the supremum over the linear combination of the function values in the unit ball of the RKHS. For a proof, see~\citep[][Lemma $3.9$]{Kanagawa2018}. 

Expanding the RKHS norm in the right hand side through an algebraic manipulation, we get
\begin{eqnarray}\nn
&&\hspace{-2em}\bigg|\bigg| k(.,x) - \sum_{i=1}^n \zeta_i(x) k(.,x_i) \bigg|\bigg|_{H_k}^2\\\nn 
&=& k(x,x) - 2\sum_{i=1}^n\zeta_i(x)k(x,x_i) + \sum_{i=1}^n\sum_{j=1}^n \zeta_i(x)\zeta_j(x)k(x_i,x_j)\\\nn
&=& k(x,x) - 2\sum_{i=1}^n \zeta_i(x)k(x,x_i) +
(Z_n(x))^{\TP}k(X_n,X_n)Z_n(x)\\\nn
&=& k(x,x) - 2 (k(x,X_n))^{\TP}(k(X_n,X_n)+\lambda^2I_n)^{-1}k(x,X_n)\\\nn
&&~~~+(k(x,X_n))^{\TP}(k(X_n,X_n)+\lambda^2I_n)^{-1}k(X_n,X_n)(k(X_n,X_n)+\lambda^2I_n)^{-1}k(x,X_n)\\\nn
&=&k(x,x) - 2 k(x,X_n)^{\TP}(k(X_n,X_n)+\lambda^2I_n)^{-1}k(x,X_n)^ \\\nn
&&~~~+k(x,X_n)^{\TP}(k(X_n,X_n)+\lambda^2I_n)^{-1}
(k(X_n,X_n)+\lambda^2I_n-\lambda^2I_n)(k(X_n,X_n)+\lambda^2I_n)^{-1}
k(x,X_n)\\\nn
&=& k(x,x) - 2 k(x,X_n)^{\TP}(k(X_n,X_n)+\lambda^2I_n)^{-1}k(x,X_n) \\\nn
&&~~~+k(x,X_n)^{\TP}(k(X_n,X_n)+\lambda^2I_n)^{-1} 
k(x,X_n)- \lambda^2 k(x,X_n)^{\TP}(k_{X_n,X_n}+\lambda^2I_n)^{-2} 
k(x,X_n)\\\nn
&=& k(x,x) - (k(X_n,X_n))^{\TP}(k(X_n,X_n)+\lambda^2I_n)^{-1}k(X_n,X_n)  - \lambda^2 k(x,X_n)^{\TP}(k(X_n,X_n)+\lambda^2I_n)^{-2} 
k(x,X_n)^{\TP}\\\nn
&=& \sigma_n^2(x) - \lambda^2 (Z_n(x))^{\TP}Z_n(x)\\\nn
&=& \sigma_n^2(x) - \lambda^2\bigg|\bigg|Z_n(x)\bigg|\bigg|^2.
\end{eqnarray}

The first equation uses the reproducing property of the RKHS. The second equation results from expressing the series in the vector product form. The third equation follows from the definition of $Z_n(x)$. The fourth and fifth equations follow from adding and subtracting a $\lambda^2 I_n$ term to the covariance matrix and some algebraic calculation. Sixth equation uses the closed form expression for the posterior variance of GP models and the definition of $Z_n(x)$.

Rearranging and combining with~\eqref{xx1}, we arrive at 
\begin{eqnarray}\nn
\sigma_n^2(x) = \sup_{f: ||f||_{H_k}\le 1}\left(f(x) - Z^{\TP}_n(x) F_n\right)^2 +\lambda^2 \bigg|\bigg|Z_n(x)\bigg|\bigg|^2.
\end{eqnarray}

\section{Proof of Theorems \textcolor{blue}{$1$} and \textcolor{blue}{$2$}}\label{app:The12}

Recall the closed form expression for the posterior mean of GP models $\mu_n(x) = Z^{\TP}_n(x)Y_n$.
We can expand the prediction error in terms of prediction error due to noise-free observations and the effect of noise as follows
\begin{eqnarray}\nn
f(x) - \mu_n(x) &=& f(x) - Z^{\TP}_n(x)Y_n \\\label{expan}
&=&f(x) - Z^{\TP}_n(x)F_n -Z^{\TP}_n(x)E_n.
\end{eqnarray}

We now use Proposition~\textcolor{blue}{$1$} to bound both terms. 

\textbf{Prediction error due to noise free observations} can be simply bounded by $B\sigma_n$ as a direct result of Proposition~1. Specifically let $\tilde{f}(.) = f(.)/B$ so that $||\tilde{f}||_{H_k}\le 1$. Also, let $\tilde{F}_n = [\tilde{f}(x_1),\tilde{f}(x_2), \dots,\tilde{f}(x_n) ]^{\TP}$.
\begin{eqnarray}\nn
|f(x) - Z^{\TP}_n(x)F_n| &=& B |\tilde{f}(x) - Z^{\TP}_n(x)\tilde{F}_n|\\\label{PEB1}
&\le& B\sigma_n(x),
\end{eqnarray}

where the inequality follows from Proposition~\tcb{$1$} and $||\tilde{f}||_{H_k}\le 1$.

We now proceed using Assumption~\tcb{2} to prove Theorem~\tcb{1}.

\textbf{The effect of noise} is bounded using the sub-Gaussianity assumption. In particular, we show that $Z^{\TP}_n(x)E_n$ is a sub-Gaussian random variable whose moment generating function is bounded by that of a Gaussian random variable with variance $\frac{R^2\sigma^2_n(x)}{\lambda^2}$. 
\begin{eqnarray}\nn
\E\bigg[\exp(Z^{\TP}_n(x)E_n)\bigg] &=& \E\left[\exp\left(\sum_{i=1}^n\zeta_i(x)\epsilon_i\right)\right]\\\nn
&=& \prod_{i=1}^n \exp(\zeta_i(x)\epsilon_i)\\\nn
&\le&  \prod_{i=1}^n \exp(\frac{R^2(\zeta_i(x))^2}{2})\\\nn
&=&\exp\left(\frac{R^2\sum_{i=1}^n(\zeta_i(x))^2}{2}\right)\\\nn
&=&\exp\left(\frac{R^2||Z_n(x)||^2}{2}\right)\\\nn
&\le&\exp\left( \frac{R^2\sigma^2_n(x)}{2\lambda^2} \right).
\end{eqnarray}
where the second equation is a result of independence of $\zeta_i(x)\epsilon_i$ terms that follows from the assumptions of i.i.d. noise terms and $X_n$ being independent of $E_n$. 
The first inequality holds by Assumption~\tcb{2}. 
We utilize Proposition~\tcb{1} to conclude that $\|Z_n(x)\|^2\le \frac{\sigma^2_n(x)}{\lambda^2}$ which results in the second inequality. Thus, using Chernoff-Hoeffding inequality~\citep{antonini2008convergence}, 
\begin{eqnarray}\nn
Z_n(x)E_n &\ge&- \frac{\sigma_n(x)R}{\lambda}\sqrt{2\log(\frac{1}{\delta})}~~~\text{w.p. at least}~1-\delta,\\\label{CH2}
Z_n(x)E_n &\le& \frac{\sigma_n(x)R}{\lambda}\sqrt{2\log(\frac{1}{\delta})}~~~\text{w.p. at least}~1-\delta.
\end{eqnarray}

Putting together~\eqref{expan},~\eqref{PEB1} and~\eqref{CH2}, Theorem~\tcb{$1$} is proven. 

We now move to the proof of Theorem~\tcb{$2$}.
For the simplicity of the notation let us use 
\begin{eqnarray}\label{eq:tau}
\tau &=& \|Z_n(x)\|\sqrt{2(\xi_0\vee\frac{2\log(1/\delta)}{h_0^2})\log(\frac{1}{\delta})},\\\label{eq:xi}
\xi &=& \xi_0\vee\frac{2\log(1/\delta)}{h_0^2}.
\end{eqnarray}
We have, for $\theta = \frac{\tau}{\xi||Z_n(x)||^2}$,

\begin{eqnarray}\nn
\Pr[Z^{\TP}_n(x)E_n\ge \tau] &=& \Pr\bigg[\exp(\theta Z^{\TP}_n(x)E_n)\ge \exp(\theta\tau)\bigg]\\\nn
&\le& \exp(-\theta \tau)\E\bigg[\exp(\theta Z^{\TP}_n(x)E_n)\bigg]\\\nn
&=& \exp(-\theta \tau)\E\left[\exp\left(\sum_{i=1}^n\theta\zeta_i(x)\epsilon_i \right)\right]\\\nn
&=& \exp(-\theta \tau) \prod_{i=1}^n\E\bigg[\exp(\theta\zeta_i(x)\epsilon_i)\bigg]\\\nn
&\le& \exp(-\theta \tau) \prod_{i=1}^n \exp\left(\frac{1}{2}\xi_0\theta^2(\zeta_i(x))^2\right)\\\nn
&=& \exp\left(\frac{1}{2}\xi_0\theta^2||Z_n(x)||^2 - \theta\tau\right)\\\nn
&=& \exp\left(
\frac{\xi_0\tau^2}{2\xi^2||Z_n(x)||^2} -
\frac{\tau^2}{\xi||Z_n(x)||^2}
\right)\\\nn
&\le& \exp(-\frac{\tau^2}{2\xi||Z_n(x)||^2})\\\label{LTCh}
&=& \delta.
\end{eqnarray}
The first line is obtained since $\exp(\theta z)$ in an increasing function in $z$. The first inequality amounts for an application of Markov inequality. The fourth line is a result of independence of $\zeta_i(x)\epsilon_i$ terms that follows from the assumptions of i.i.d. noise terms and $X_n$ being independent of $E_n$. The second inequality holds by definition of light-tailed distributions. Notice that the careful choice of $\tau$ and $\theta$ ensures $\theta\zeta_i(x)\le h_0$, which will be validated next. The seventh line is obtained by replacing the value of $\theta$. The last inequality is obtained by $\xi_0\le \xi$. The last line is resulted from replacing the value of $\tau$ from~\eqref{eq:tau}.

It remains to validate $\theta\zeta_i(x)\le h_0$.
\begin{eqnarray}\nn
\theta\zeta_i(x) &=& \frac{\tau}{\xi||Z_n(x)||^2}\zeta_i(x)\\\nn
&=&\frac{\sqrt{2\log(\frac{1}{\delta})}\zeta_i(x)}{\sqrt{\xi}||Z_n(x)||}\\\nn
&\le& h_0\frac{\zeta_i(x)}{||Z_n(x)||}\\\nn
&\le& h_0,
\end{eqnarray}
where we replace 
$\theta = \frac{\tau}{\xi||Z_n||^2}$, and the values of $\tau$ and $\xi$ from~\eqref{eq:tau} and~\eqref{eq:xi}, respectively. For the first inequality, notice that $\frac{2\log(1/\delta)}{h_0^2}\le \xi$ from the definition of $\xi$~\eqref{eq:xi}. For the second inequality notice that $\zeta_i(x)\le ||Z_n(x)||$.

We thus proved 
\begin{eqnarray}\label{eq7lt}
Z_n(x)E_n \le \tau,~~~\text{w.p. at least}~1-\delta.
\end{eqnarray}
Similarly, we can prove 
\begin{eqnarray}\label{eq8lt}
Z_n(x)E_n \ge -\tau,~~~\text{w.p. at least}~1-\delta.
\end{eqnarray}

Replacing $||Z_n(z)||\le \frac{R}{\lambda}\sigma_n(x)$ from Proposition~\tcb{$1$} in the value of $\tau$~\eqref{eq:tau}, and combining~\eqref{eq7lt} and~\eqref{eq8lt} with~\eqref{expan} and \eqref{PEB1}, Theorem~\tcb{$2$} is proven.

\section{Proof of Theorem \tcb{$3$}}\label{app:The3}

The MVR algorithm selects the points with the highest variance first. Thus, $\forall x\in \Xc$,
\begin{eqnarray}\label{eq9}
\sigma^2_{n-1}(x)\le \sigma^2_{n-1}(x_n).
\end{eqnarray}
By definition of conditional variance of normal distributions and due to positive definiteness of covariance matrix, conditioning on a larger set of points reduces the variance. Specifically,
we have, for all $x\in \Xc$ and $\forall n \le N$,
\begin{eqnarray}\label{eq10}
\sigma^2_{N}(x)\le \sigma^2_{n-1}(x).
\end{eqnarray}

Combining~\eqref{eq9} and~\eqref{eq10}, we have, $\forall x\in\Xc$ and $\forall n\le N$,
\begin{eqnarray}\nn
\sigma^2_{N}(x)\le \sigma^2_{n-1}(x_n).
\end{eqnarray}
Averaging both sides over $n$ (from $1$ to $N$), we have
\begin{eqnarray}\label{sigub1}
\sigma^2_{N}(x) \le \frac{1}{N}\sum_{n=1}^N \sigma^2_{n-1}(x_n).
\end{eqnarray}

We now use the following lemma to bound $\sigma^2_N(x)$.

\begin{lemma}\label{DnIn}
Recall $\Ic(Y_n; \hat{f})=\frac{1}{2}\log\det(I_n+\frac{1}{\lambda^2}k(X_n,X_n))$. For the cumulative conditional variance of the GP model,
we have
\begin{eqnarray}\nn
\sum_{n=1}^N\sigma_{n-1}^2(x_n)\le \frac{2}{\log(1+\frac{1}{\lambda^2})}\Ic(Y_n;\hat{f}).
\end{eqnarray}
\end{lemma}
%\begin{proof}[Proof of Lemma~\ref{DnIn}]
A proof can be found in \cite{srinivas2010gaussian}. 

%\end{proof}

We thus have, for all $x\in \Xc$,
\begin{eqnarray}\nn
\sigma^2_{N}(x) &\le&
\frac{2\Ic(Y_n;\hat{f})}{\log(1+\frac{1}{\lambda^2})N}\\\label{sigub2}
&\le&\frac{2\gamma_N}{\log(1+\frac{1}{\lambda^2})N},
\end{eqnarray}
where $\gamma_N$ is the maximal information gain defined in Sec.~\tcb{2.4}. 

Let $B_0(\delta) = B+\sqrt{N}\beta(2\delta/N)$. At the end of this section, in Lemma~\ref{Lemma:RKHSnormMu}, we prove that  
\begin{eqnarray}
\|\mu_N\|_{H_k} \le B_0(\delta),~\text{w.p. at least}~1-\delta. 
\end{eqnarray}

Notice that $\mu_n$ is a random function due to the randomness in noise. Let us define the event $\Ec =\{\|\mu_N\|_{H_k} \le B_0(\delta/3)\}$. We have $\Pr[\Ec]\ge 1-\frac{\delta}{3}$. 

Under event $\Ec$, we use Assumption~\tcb{4} on the existence of a discretization $\Dc_N(\delta)$ of $\Xc$ such that $f(x) - f([x]_N)\le \frac{1}{\sqrt{N}}$, $\mu_N(x) -\mu_N([x]_N)\le \frac{1}{\sqrt{N}}$, and $|\Dc_N(\delta)|\le CB^d_0(\delta/3)N^{d/2}$. Notice that we do not need to actually create this discretization. We only use its existence to handle the analysis in a continuous space using a probability union bound based on this discretization. 

For a fixed $x\in \Dc_N$, from the confidence bounds for GP models proven in Theorems~\tcb{$1$} and~\tcb{$2$}, we have
\begin{eqnarray}\nn
% f(x)  &\le&  \mu_n(x) + \beta_n^{\frac{\delta}{Cn^{d/2}}}\sigma_n(x)~~~\text{w.p.}~1-\frac{\delta}{C2^n}\\
f(x)  &\ge& \mu_n(x) - (B+\beta({\frac{\delta}{3|\Dc_N(\delta)|}}))\sigma_n(x),~~~\text{w.p. at least}~1-\frac{\delta}{3|\Dc_N(\delta)|}.
\end{eqnarray}
 
Using a probability union bound, we have, $\forall x\in \Dc_N(\delta)$
\begin{eqnarray}\label{un1}
% f(x)  &\le&  \mu_n(x) + \beta_n^{\frac{\delta}{Cn ^{d/2}}}\sigma_n(x)~~~\text{w.p.}~1-\delta\\
f(x)  &\ge& \mu_n(x) - (B+ \beta({\frac{\delta}{3|\Dc_N(\delta)|}}))\sigma_n(x),~~~\text{w.p. at least}~1-\frac{\delta}{3}.
\end{eqnarray}

% In addition, from Assumption~\textcolor{blue}{4}, we have, $\forall x\in \Xc$
% \begin{eqnarray}
% |f(x) - f([x]_N)| \le \frac{1}{\sqrt{N}}
% \end{eqnarray}

% Let $z^*  =\text{argmax}_{x\in \Dc_N}f(x)$ and $\hat{z}^*_N = \text{argmax}_{x\in \Dc_N}\mu_N(x)$. 
We thus have, under event $\Ec$,
\begin{eqnarray}\nn
f(x^*) - f(\hat{x}^*_N) &=&
f(x^*)  - f([\hat{x}^*_N]_N) + f([\hat{x}^*_N]_N) - f(\hat{x}^*_N)\\\nn
&\le& f(x^*) - f([\hat{x}^*_N]_N) + \frac{1}{\sqrt{N}}\\\nn
&\le& f(x^*) - \mu_N(x^*) + \mu_N(\hat{x}^*_N) - f([\hat{x}^*_N]_N) + \frac{1}{\sqrt{N}}\\\nn
&=&f(x^*) - \mu_N(x^*) +
\mu_N(\hat{x}^*_N) - \mu_N([\hat{x}^*_N]_N)
+ \mu_N([\hat{x}^*_N]_N)
 - f([\hat{x}^*_N]_N) + \frac{1}{\sqrt{N}}\\\nn
&\le&f(x^*) - \mu_N(x^*)
+ \mu_N([\hat{x}^*_N]_N)
 - f([\hat{x}^*_N]_N) + \frac{2}{\sqrt{N}}.\\\nn
\end{eqnarray}

The first inequality comes from Assumption~\tcb{$4$} on discretization $\Dc_N(\delta)$ and $f$. The second inequality comes from the definition of MVR which ensures $\mu_N(\hat{x}^*_N)\ge \mu_N(x)$, for all $x\in\Xc$. For the last inequality, we use Assumption~\tcb{$4$} on discretization $\Dc_N(\delta)$ and $\mu_N$. Notice that under event $\Ec$, the posterior mean of the GP model belongs to the same RKHS with its norm bounded by $B_0(\delta/3)$. 

Thus, assuming that the inequality given in \eqref{un1}, the confidence interval for $f(x^*)$ with $1-\delta/3$ confidence, and $\Ec$, all three hold true (notice that each of these three events hold true with probability at least $1-\frac{\delta}{3}$), using a probability union bound, we have 

\begin{eqnarray}
 f(x^*) - f(\hat{x}^*_N) &\le& (B+\beta({\frac{\delta}{3}}))\sigma_N(x^*) +(B+\beta({\frac{\delta}{3|\Dc_N(\delta)|}})\sigma_N([\hat{x}^*_N]_N)\\\nn
 &&~~~~~+ \frac{2}{\sqrt{N}},~~~ \text{w.p. at least}~1-\delta.
\end{eqnarray}

Using~\eqref{sigub2} to bound $\sigma_N(x^*)$ and $\sigma_N([\hat{x}^*_N]_N)$, and replacing $|\Dc_N(\delta)|$ with its upper bound, we get
\begin{eqnarray}\nn
f(x^*) - f(\hat{x}^*_N) &\le& \sqrt{\frac{2\gamma_N}{\log(1+\frac{1}{\lambda^2})N}}
\left(
2B
+\beta(\frac{\delta}{3}) 
+ \beta(\frac{\delta}{
3C(B+\sqrt{N}\beta(2\delta/3N))^dN^{d/2}
})
\right) \\
&&~~~~~
+ \frac{2}{\sqrt{N}},~\text{w.p. at least}~1-\delta,
\end{eqnarray}
which completes the proof. 

% Under Assumptions~1 and 3,
% \begin{eqnarray}
% r_N^{\text{MVR}} \le \sqrt{\frac{2\gamma_N}{\log(1+\frac{1}{\lambda^2})N}}\left(2B + \frac{R}{\lambda}\left(\sqrt{2\log(\frac{2}{\delta})} +\sqrt{\frac{d}{2}\log(N) + \log(\frac{2C}{\delta})}\right)
% \right)
% \end{eqnarray}

We now prove a high probability upper bound on $\|\mu_n\|_{H_k}$. 

Let us first formally state the equivalence of the posterior mean in GP models and the regressor in kernel ridge regression.

\begin{lemma}\label{KRR}
Conditioned on a set of noisy observation $\Hc_n$ from $f$, recall the expression for the posterior mean of the GP model $\mu_n(x) = Z^{\TP}_n(x)Y_n$. We have the following equality 
\begin{eqnarray}\label{Lemma:equiv}
\mu_n = \text{argmin}_{g\in H_k} \left(\lambda^2||g||^2_{H_k} + \sum_{i=1}^n (g(x_i)-y_i)^2\right).
\end{eqnarray}
\end{lemma}
For a proof, see~\citep[][Theorem 3.4]{Kanagawa2018}. Lemma~\ref{KRR} establishes the equivalence of the posterior mean in GP models and the regressor in kernel ridge regression. It indicates that the posterior mean of GP models is a mean squared error estimator, regularized by the RKHS norm, where $\lambda^2$ is the regularization parameter. We use this lemma to show that the posterior mean of the GP model with high probability lives in the same RKHS as $f$.

\begin{lemma}\label{Lemma:RKHSnormMu}
Conditioned on a set of noisy observation $\Hc_n$ from $f$ with $\|f\|_{H_k}\le B$, the RKHS norm of the posterior mean of the GP model $\mu_n(x) = Z^{\TP}_n(x)Y_n$ satisfies the following
\begin{eqnarray}
\|\mu_n\|_{H_k} \le B +  \sqrt{n}\beta(2\delta/n),~\text{w.p. at least}~1-\delta,
\end{eqnarray}
where $\beta(\delta) = \frac{R}{\lambda}\sqrt{2\log(\frac{1}{\delta})} $ under Assumption \tcb{2}, and $\beta(\delta) = \frac{1}{\lambda}\sqrt{2(\xi_0\vee\frac{2\log(1/\delta)}{h_0^2})\log(\frac{1}{\delta})}$ under Assumption \tcb{3}.

\end{lemma}

\paragraph{Proof of Lemma~\ref{Lemma:RKHSnormMu}:}

We have
\begin{eqnarray}\nn
\|\mu_n\|_{H_k} &=&
\|Z_n^{\TP}(.)F_n + Z_n^{\TP}(.)E_n\|_{H_k} \\\label{knkn}
&\le& \|Z_n^{\TP}(.)F_n \|_{H_k}  + \| Z_n^{\TP}(.)E_n\|_{H_k}.
\end{eqnarray}

From Lemma~\ref{KRR}, we have
\begin{eqnarray}\nn
\lambda^2\|Z_n^{\TP}(.)F_n \|^2_{H_k} + \sum_{i=1}^n(Z_n^{\TP}(x_i)F_n-f(x_i))^2 \le  
\lambda^2\|f \|^2_{H_k} + \sum_{i=1}^n(f(x_i)-f(x_i))^2
\end{eqnarray}
Thus,
\begin{eqnarray}
\|Z_n^{\TP}(.)F_n \|_{H_k}  \le  
\|f \|_{H_k},
\end{eqnarray}
where $\|f\|_{H_k}\le B$.
It thus remains to bound the second term on the right hand side of~\eqref{knkn}.
\begin{eqnarray}\nn
\| Z_n^{\TP}(.)E_n\|^2_{H_k} &=& \|k^{\TP}(x,X_n)\left(k(X_n,X_n)+\lambda^2 I_n\right)^{-1}E_n\|^2_{H_k}\\\nn
&=& E_n^{\TP}\left(k(X_n,X_n)+\lambda^2 I_n\right)^{-1}k(X_n,X_n)
\left(k(X_n,X_n)+\lambda^2 I_n\right)^{-1}E_n\\\nn
&=& E_n^{\TP}\left(k(X_n,X_n)+\lambda^2 I_n\right)^{-1}\left(k(X_n,X_n)+\lambda^2 I_n\right)
\left(k(X_n,X_n)+\lambda^2 I_n\right)^{-1}E_n\\\nn
&&\hspace{2em}-\lambda^2 E_n^{\TP}\left(k(X_n,X_n)+\lambda^2 I_n\right)^{-2}E_n\\\nn
&\le& E_n^{\TP}
\left(k(X_n,X_n)+\lambda^2 I_n\right)^{-1}E_n\\\nn
&\le&\frac{1}{\lambda^2}\|E_n\|^2_{l_2},
\end{eqnarray}
where
for the second line we used the reproducing property of the RKHS, 
for the first inequality we used 
positive definiteness of $\left(k(X_n,X_n)+\lambda^2 I_n\right)^{-2}$ that is a result of positive definiteness of $k(X_n,X_n)$, and for the last inequality we used positive definiteness of $k(X_n,X_n)$.

Under Assumption~\tcb{$2$}, as a result of Chernoff-Hoeffding inequality,
\begin{eqnarray}\nn
\epsilon_i^2 \le 2R^2\log(\frac{1}{2\delta'}),~\text{w.p. at least}~ 1-\delta'. 
\end{eqnarray}
Using a probability union bound over $i=1,2,\dots,n$, with $\delta'=\frac{\delta}{n}$,
\begin{eqnarray}
\frac{1}{\lambda^2}\|E_n\|^2_{l_2} \le \frac{2nR^2}{\lambda^2}\log(\frac{n}{2\delta}),~\text{w.p. at least}~ 1-\delta.
\end{eqnarray}

Under Assumption~\tcb{$3$}, as a result of~\eqref{LTCh} (with $n=1$, and $Z_n=1$),
\begin{eqnarray}\nn
\epsilon_i^2 \le 2(\xi_0\vee\frac{2\log(1/2\delta')}{h_0^2})\log(\frac{1}{2\delta'}),~\text{w.p. at least}~ 1-\delta'.
\end{eqnarray}

Using a probability union bound over $i=1,2,\dots,n$, with $\delta'=\frac{\delta}{n}$,
\begin{eqnarray}
\frac{1}{\lambda^2}\|E_n\|^2_{l_2} \le \frac{2n}{\lambda^2}(\xi_0\vee\frac{2\log(n/2\delta)}{h_0^2})\log(\frac{n}{2\delta}),~\text{w.p. at least}~ 1-\delta.
\end{eqnarray}

Combining the bounds on the both terms on the right hand side of~\eqref{knkn}, the lemma is proven.

\section{Proof of Corollary 1}\label{app:cor1}

We use Theorem~\tcb{3} to derive a bound on the expected regret of MVR. 

First, notice that $|f(x)|\le {k}_0B$ where $k^2_0=\max_{x\in \Xc}k(x,x)$, which can be proven using the reproducing property of the RKHS.
\begin{eqnarray}\nn
|f(x)| &=& |\langle f(.), k(.,x)\rangle_{H_k}|\\\nn
&\le& ||f||_{H_k}||k(.,x)||_{H_k}\\\nn
&=& ||f||_{H_k}\sqrt{k(x,x)}\\\nn
&\le& k_0 B.
\end{eqnarray}

So, we have $\max_{x\in\Xc} f(x^*) - f(x)\le 2k_0B$. Let $\Ec$ denote the even that $r^{\text{MVR}}_{N}\le
\bar{r}
$, where
\begin{eqnarray}\nn
\bar{r}= \sqrt{\frac{2\gamma_N}{\log(1+\frac{1}{\lambda^2})N}}
\left(
2B
+\beta(\frac{1}{3\sqrt{N}}) 
+ \beta\left(\frac{1}{
3C\sqrt{N}\left(B+\sqrt{N}\beta(2/3N\sqrt{N})\right)^dN^{d/2}
}\right)
\right) 
+ \frac{2}{\sqrt{N}}
\end{eqnarray}
is the upper bound on regret given in Theorem~\tcb{$3$} with $\delta=\frac{1}{\sqrt{N}}$.
From Theorem~\tcb{3}, we have $\Pr[\Ec]\ge 1-\frac{1}{\sqrt{N}}$. 

Using the law of total expectation, we have
\begin{eqnarray}\nn
\E[r^{\text{MVR}}_{N}] &=& \E\left[r^{\text{MVR}}_{N}|\Ec\right]\Pr[\Ec] +\E\left[r^{\text{MVR}}_{N}|\bar{\Ec}\right]\Pr[\bar{\Ec}]
\\\nn
&\le& \bar{r} + \frac{2k_0B}{\sqrt{N}}\\\nn
&=& \Oc\left(\sqrt{\frac{\gamma_N}{N}}\beta(N^{d+\frac{1}{2}}) \right). 
\end{eqnarray}

Under Assumption~\tcb{$2$}
\begin{eqnarray}
\E[r^{\text{MVR}}_{N}] = \Oc\left(\sqrt{\frac{\gamma_N}{N}\log(N^{d+\frac{1}{2}})}\right).
\end{eqnarray}

Under Assumption~\tcb{$3$}
\begin{eqnarray}
\E[r^{\text{MVR}}_{N}] = \Oc\left(\sqrt{\frac{\gamma_N}{N}}\log(N^{d+\frac{1}{2}})\right).
\end{eqnarray}

For SE kernel, $\gamma_N =\Oc\left( \log^{d+1}(N)\right)$~\citep{srinivas2010gaussian}. Selecting $N\propto (\frac{1}{\epsilon})^2\log^{d+2}(\frac{1}{\epsilon})$ and $N\propto (\frac{1}{\epsilon})^2\log^{d+3}(\frac{1}{\epsilon})$, with proper constants, under Assumptions~\tcb{$2$} and~\tcb{$3$}, respectively, results in $\E[r^{\text{MVR}}_{N}]\le\epsilon$. 

In the case of Mat{\'e}rn kernel, $\gamma_N = \Oc\left(N^{\frac{d}{2\nu+d}}(\log(N))^{\frac{2\nu}{2\nu+d}}\right)$~\citep{vakili2020information}.
Selecting $N\propto (\frac{1}{\epsilon})^{2+\frac{d}{\nu}}
(\log(\frac{1}{\epsilon}))^{\frac{4\nu+d}{2\nu}}$ and $N\propto (\frac{1}{\epsilon})^{2+\frac{d}{\nu}}
(\log(\frac{1}{\epsilon}))^{\frac{6\nu+2d}{2\nu}}$, with proper constants, under Assumptions~\tcb{$2$} and~\tcb{$3$}, respectively, results in $\E[r^{\text{MVR}}_{N}]\le\epsilon$. 

Finding the exact constants requires solving a non-linear equation involving $\log$ function which is a tedious task. 

Noticing that $\E[r^{\text{MVR}}_{n}]$ is a decreasing function in $n$ completes the proof.

\section{Supplemental Material on the Experiments}\label{app:exp}

In this section, we provide further details on the experiments. We also provide additional experiments on two commonly used benchmark functions for Bayesian optimization.

\subsection{Additional Experiments}

\begin{figure*}[ht]
\centering
    
    %%%
    \subfloat[SE, Gaussian Noise]{ \centering \includegraphics[width=0.30\columnwidth]{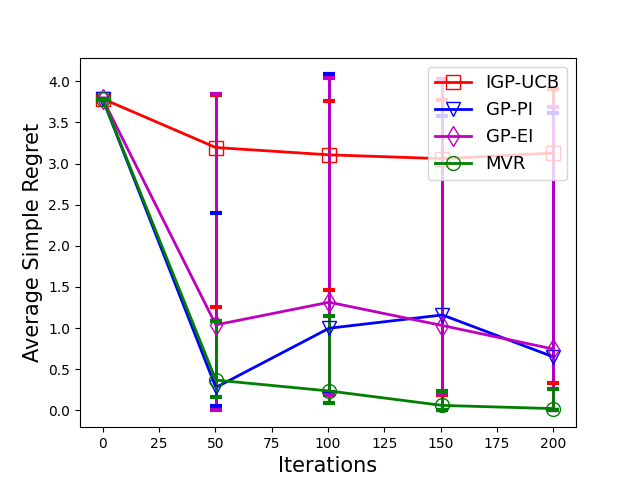}}
    %%%
    \subfloat[Mat{\'e}rn, Gaussian Noise]{ \centering \includegraphics[width=0.30\columnwidth]{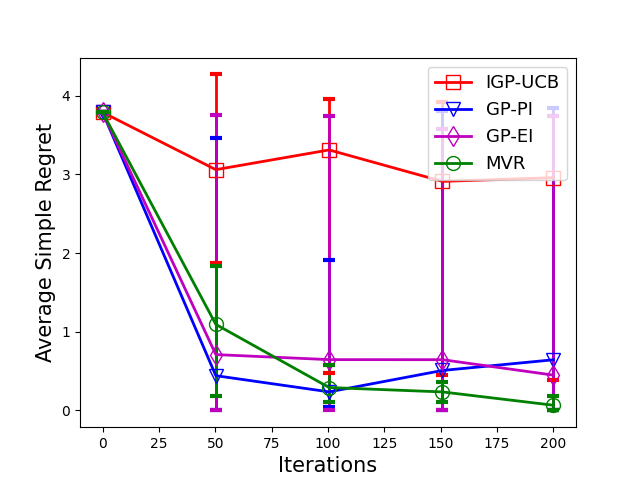}}

   %%%
    \subfloat[SE, Laplace Noise]{ \centering \includegraphics[width=0.30\columnwidth]{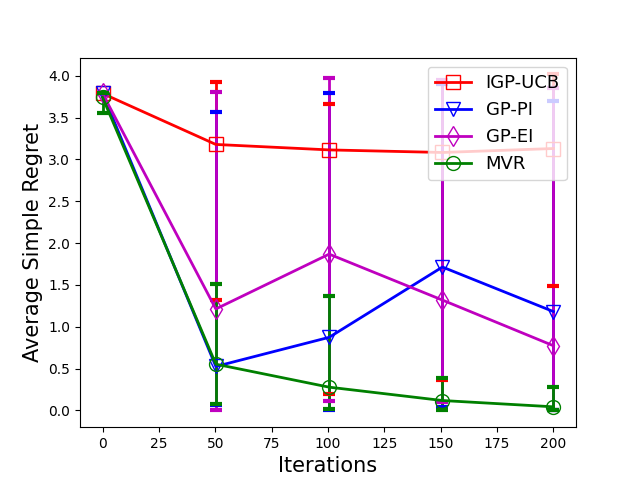}}
    %%%
    \subfloat[Mat{\'e}rn, Laplace Noise]{ \centering \includegraphics[width=0.30\columnwidth]{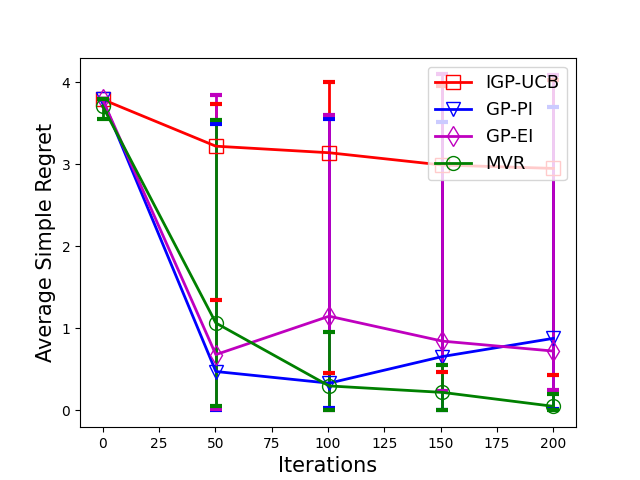}}
\caption{Comparison of the simple regret performance of Bayesian optimization algorithms on Hartman3 test function.}
\label{Fig2}
\end{figure*}

\begin{figure*}[ht]
\centering
    
    %%%
    \subfloat[SE, Gaussian Noise]{ \centering \includegraphics[width=0.30\columnwidth]{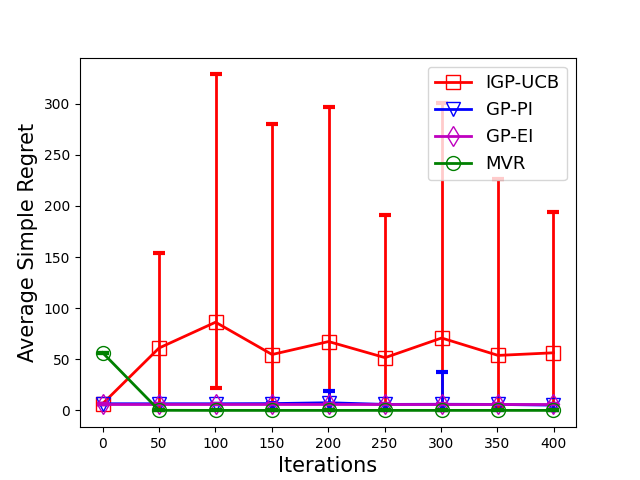}}
    %%%
    \subfloat[Mat{\'e}rn, Gaussian Noise]{ \centering \includegraphics[width=0.30\columnwidth]{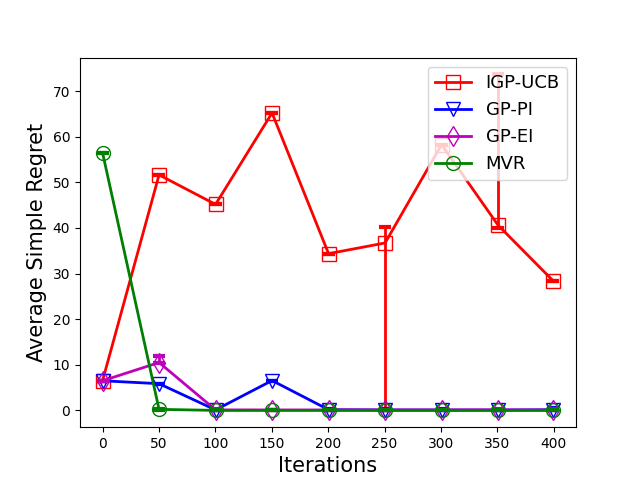}}

   %%%
    \subfloat[SE, Laplace Noise]{ \centering \includegraphics[width=0.30\columnwidth]{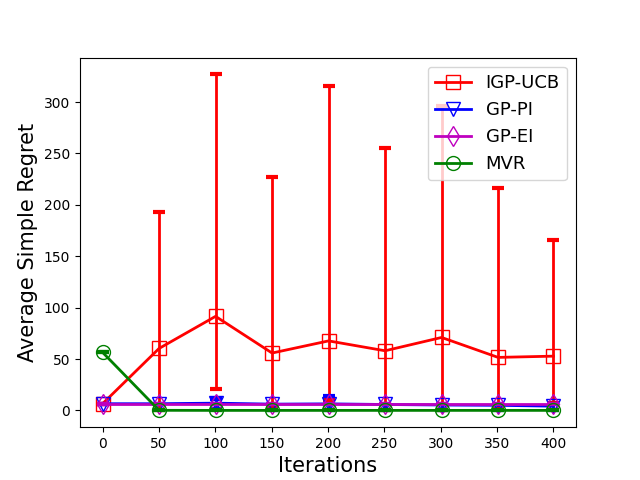}}
    %%%
    \subfloat[Mat{\'e}rn, Laplace Noise]{ \centering \includegraphics[width=0.30\columnwidth]{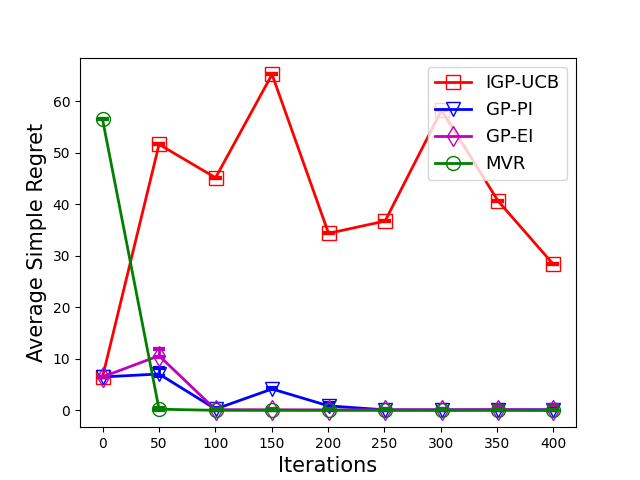}}
\caption{Comparison of the simple regret performance of Bayesian optimization algorithms on Rosenbrock test function.}
\label{Fig3}
\end{figure*}

In Section~\tcb{$5$}, we provided experiments on the comparison of the simple regret performance of Bayesian optimization algorithms on synthetically generated functions in RKHS. 
In this section, we consider two commonly used benchmark functions for Bayesian optimization: Hartman3 and Rosenbrock as presented in~\cite{azimi2012hybrid, Picheny2013}. The parameters of the kernels, noise and $\lambda$ are set exactly as described in Section \tcb{$5$}. We plot the average simple regret for all four learning algorithms considered in Section \tcb{$5$}, over $50$ independent experiments, with Hartman3 test function in Figure~\ref{Fig2}, and with Rosenbrock test function in Figure~\ref{Fig3}. The details on the source code is provided in the next section. The data used for generating the figures is provided in the supplementary material.

\subsection{Additional Details on the Experiments}

In the paper, we have provided a complete theoretical analysis of sample complexity. Here, we briefly mention the computational complexity of the algorithms. There are two computational bottlenecks in implementing Bayesian optimization algorithms. First bottleneck is the update of the GP model based on past observations which requires an $\Oc(n^3)$ computation at time $n$, due to the matrix inversion, $(k(X_n,X_n)+\lambda^2I_n)^{-1}$, step. 
Sparse approximations of matrix inversion~\cite{Calandriello2019Adaptive} or sparse variational models~\cite{Titsias2009Variational, Hensman2013, Vakili2020Scalable} can reduce the computational complexity from $\Oc(n^3)$ to $\Oc(n)$, however at the price of introducing an approximation error.  
Second bottleneck is the selection of the observation point based on the \emph{acquisition functions} which are summarized next for each algorithm.
\begin{itemize}
    \item IGP-UCB: $\mu_n(x)+\beta_n^{\delta}\sigma_n(x)$ where $\beta_n^{\delta} = \left(B+R\sqrt{2(\gamma_n+1+\log(\frac{1}{\delta}))}\right)$.
    \item GP-PI: $\Pr[f(x)\ge \mu^+ +\alpha] = \Phi\left(\frac{\mu_n(x)-\mu^+-\alpha}{\sigma_n(x)}\right)$, where $\mu^+=\max_{i<n}\mu_{i-1}(x_i)$, $\alpha>0$ is a user selected hyper-parameter (set to $0.01$ in our experiments as suggested in~\cite{hoffman2011portfolio}), and $\Phi$ is the cumulative density function of the standard normal distribution. 
    \item GP-EI: $\kappa\Phi(\frac{\kappa}{\sigma_n(x)}) + \sigma_n(x)\phi(\frac{\kappa}{\sigma_n(x)})$, where $\kappa = \mu_n(x) - \mu^+-\alpha$, and $\phi$ and $\Phi$ denote the probability density function and cumulative density function of the standard normal distribution, respectively. The parameters $\mu^+$ and $\alpha$ are set similar to GP-PI, following~\cite{hoffman2011portfolio}.
\end{itemize}

The standard approach in finding the maximizer of the acquisition function is to evaluate it on a grid discretizing the search space~\citep{Chowdhury2017bandit}. For a grid of size $M$, this requires $O(Mn)$ computations at time $n$. We have used the same discretization for all algorithms.

A practical idea to improve the computational cost in implementing Bayesian optimization algorithms is to use an off-the-shelf optimizer to solve the optimization of the acquisition function at each iteration (instead of using a grid). This method, although can lead to significant gains in computational complexity, invalidates the existing regret bounds, due to lack of guarantees for an accurate optimization of the acquisition function (that is often non-convex). We have thus used the discretization method following most related work with analytical regret guarantees~\citep[e.g.,][]{srinivas2010gaussian, Chowdhury2017bandit}.

\end{appendix}

\end{document}